\documentclass [twoside,11pt] {article}
\usepackage{jmlr2e}
\firstpageno{1}
\usepackage{url}

\begin{document}

\title{
Universal Algorithm for Online Trading Based on the Method of Calibration
}

\author{\name Vladimir V. V'yugin \email vyugin@iitp.ru \\
       \addr Institute for Information Transmission Problems\\
       Russian Academy of Sciences\\
       Bol'shoi Karetnyi per. 19 \\ Moscow GSP-4, 127994, Russia
       \AND
       \name Vladimir G. Trunov \email trunov@iitp.ru \\
       \addr Institute for Information Transmission Problems\\
       Russian Academy of Sciences\\
       Bol'shoi Karetnyi per. 19 \\ Moscow GSP-4, 127994, Russia
       }

\editor{}

\maketitle

\begin{abstract}%
We present a universal method for algorithmic trading in Stock Market
which performs asymptotically at least as well as any stationary trading strategy
that computes the investment at each step using a fixed function of the
side information that belongs to a given RKHS
(Reproducing Kernel Hilbert Space). Using a universal kernel,
we extend this result for any continuous stationary strategy.
In this learning process, a trader rationally chooses his gambles
using predictions made by a randomized well-calibrated
algorithm. Our strategy is based on Dawid's notion of
calibration with more general checking rules and on
some modification of Kakade and Foster's randomized rounding algorithm
for computing the well-calibrated forecasts. We combine the method of
randomized calibration with Vovk's method of
defensive forecasting in RKHS.
Unlike in statistical theory, no stochastic assumptions are made about
the stock prices. Our empirical results on historical markets provide strong
evidence that this type of technical trading can ``beat the market''
if transaction costs are ignored.
\end{abstract}


\section{Introduction}\label{intr-1}

Predicting sequences is the key problem for machine learning, computational
finance and statistics. These predictions can serve as a base
for developing the efficient methods for playing financial games
in Stock Market.

The learning process proceeds as follows: observing a finite-state
sequence given online, a forecaster assigns a subjective estimate to
future states.

A minimal requirement for testing any prediction algorithm is that
it should be calibrated (cf.~\citealt{Daw82}).
Dawid gave an informal explanation of calibration for binary outcomes.
Let a sequence $\omega_1,\omega_2,\dots ,\omega_{n-1}$ of binary outcomes
be observed by a forecaster whose task is to give a probability $p_n$
of a future event $\omega_n=1$. In a typical example, $p_n$ is interpreted
as a probability that it will rain. Forecaster is said to be
well-calibrated if it rains as often as he leads us to expect. It should rain
about $80\%$ of the days for which $p_n=0.8$, and so on.

A more precise definition is as follows.
Let $I(p)$ denote the characteristic function of a
subinterval $I\subseteq [0,1]$, i.e., $I(p)=1$ if $p\in I$ and $I(p)=0$,
otherwise.
An infinite sequence of forecasts $p_1,p_2,\dots$ is calibrated
for an infinite binary sequence of outcomes
$\omega_1\omega_2\dots$ if for characteristic function $I(p)$
of any subinterval of $[0,1]$ the calibration error tends to zero, i.e.,
\begin{eqnarray*}\label{call-1}
\frac{1}{n}\sum_{i=1}^{n}I(p_i)(\omega_i-p_i)\to 0
\end{eqnarray*}
as $n\to\infty$.
The indicator function $I(p_i)$ determines some
``checking rule'' that selects indices $i$, where we compute
the deviation between forecasts $p_i$ and outcomes $\omega_i$.

If the weather acts adversatively, then, as shown by~\citet{Oak85}
and~\citet{Daw85}, any deterministic
forecasting algorithm will not always be calibrated.

\citet{FoV98} show that calibration is almost surely
guaranteed with a randomizing forecasting rule, i.e., where the forecasts
$p_i$ are chosen using internal randomization and the forecasts are hidden
from the weather until the weather makes its decision whether to rain or not.

The origin of the calibration algorithms is the~\citet{Bla56}
approachability theorem
but, as its drawback, the forecaster has to use linear programming to compute
the forecasts. We modify and generalize a more computationally efficient method from
~\citet{KaF2004}, where ``an almost deterministic''
randomized rounding universal forecasting algorithm is presented.
For any sequence of outcomes $\omega_1,\omega_2,\dots$ and for any precision of
rounding $\Delta>0$,
an observer can simply randomly round the deterministic forecast $p_i$ up to
$\Delta$ to a random forecast $\tilde p_i$ in order to calibrate
for this sequence with probability one:
\begin{eqnarray}\label{call-1a}
\limsup\limits_{n\to\infty}\left|\frac{1}{n}\sum_{i=1}^{n}
I(\tilde p_i)(\omega_i-\tilde p_i)\right|\le\Delta,
\end{eqnarray}
where $I(p)$ is the characteristic function of any subinterval of $[0,1]$.
This algorithm can be easily generalized
such that the calibration error tends to zero as $n\to\infty$.

Kakade and Foster and others considered a finite outcome space
and a probability distribution as the forecast. In this paper,
the outcomes $\omega_i$ are real numbers from unit interval $[0,1]$ and the
forecast $p_i$ is a single real number (which can
be an output of a random variable). This setting is closely
related to~\citet{Vov2005a} defensive forecasting approach
(see below).

In this case real valued predictions $p_i\in [0,1]$ could
be interpreted as mean values of future outcomes under
some unknown to us probability distributions in $[0,1]$. We do not know
precise form of such distributions -- we should predict only future means.

The well known applications of the method of calibration belong
to different fields of the game theory and machine learning.
Kakade and Foster proved that empirical
frequencies of play in any normal-form game with finite
strategy sets converges to a set of correlated equilibrium if
each player chooses his gamble as the best response to the well
calibrated forecasts of the gambles of other players. In series
of papers:
~\citet{TaV2005},~\citet{Vov2005a},~\citet{Vov2006},~\citet{Vov2006a},
~\citet{Vov2007}, Vovk developed the method of calibration for the
case of more general RKHS and
Banach spaces. Vovk called his method defensive forecasting
(DF). He also applied his method for recovering unknown
functional dependencies presented by arbitrary functions from
RKHS and Banach spaces. \citet{CKZV2010} show
that well-calibrated forecasts can be used to compute
predictions for the~\citet{Vov97} aggregating algorithm.
In defensive forecasting, continuous loss (gain) functions are considered.

In this paper we present a new application of the method of
calibration. We construct ``a universal'' strategy for algorithmic trading in
{\it Stock Market} which performs asymptotically at least as well as any not ``too
complex'' trading strategy $D$. Technically, we are
interested in the case where the trading strategy $D$ is
assumed to belong to a large reproducing kernel Hilbert space
(to be defined shortly) and the complexity of $D$ is measured
by its norm. Using a universal kernel, we extend this result
to any continuous stationary trading strategy. Our universal
trading strategy is represented by a discontinuous function though
it uses a randomization.

First discuss some standard financial terminology.
A trader in Stock Market uses a strategy: going long or going short,
or skip the step.
In finance, a long position in a security, such as a stock or a bond,
or equivalently to be long in a security, means that the holder of the position
owns the security and will profit if the price of the security goes up.
Short selling (also known as shorting or going short) is the practice of
selling securities or other financial instruments, with the intention of
subsequently repurchasing them (``covering'') at a lower price.

In this paper, the problem of universal sequential investment in Stock Market
with side information is studied.
We consider the method of trading called in financial industrial
applications {\it algorithmic trading} or
{\it systematic quantitative trading}, which
means rule-based automatic trading strategies, usually implemented
with computer based trading systems.

The problem of algorithmic
trading is considered in machine learning framework, where algorithms
adaptive to input data are designed and their performance is evaluated.

There are three common types of analysis for adaptive algorithms: average case
analysis which requires a statistical model of input data; worst-case
analysis which is non-informative because, for any trading algorithm,
we can present a sequence of stock prices moving in the
direction opposite to the trader's decisions;
competitive analysis which is popular
in the prediction with expert advice framework.

A non-traditional objective (in computational finance) is to
develop algorithmic trading strategies that are in some sense always
guaranteed to perform well. In competitive analysis, the
performance of an algorithm is measured to any trading
algorithm from a broad class. We only ask than an algorithm
performs well relative to the difficulty in classsifying of the input data.
Given a particular performance measure, an adaptive algorithm is
strongly competitive with a class of trading algorithms if it
achieves the maximum possible regret over all input sequences.
Unlike in statistical theory, no stochastic assumptions are
made about the stock prices.

This line of research in finance was pioneered by Cover
(see~\citealt{CoG86}, \citealt{Cov91}, \citealt{CoO96}) who designed
universal portfolio selection algorithms that can provably do well (in terms
of their total return) with respect to some adaptive online or offline
benchmark algorithms. Such algorithms are called {\it universal}.

We consider the simplest case: algorithmic trading with only stock.
Our results can be generalized for the case of several stocks
and for dynamical portfolio hedging in sense of framework proposed
by~\citet{CoO96}.

We consider a game with players: {\it Stock Market} and {\it Trader}.
At the beginning of each round $i$ {\it Trader} is shown an object
${\bf x}_i$ which contains a side information. Past prices of the stock
$S_1,\dots, S_{i-1}$ are also given for {\it Trader} (they can be considered
as a part of the side information). Using this information, {\it Trader}
announces a number $M_i$ of shares of the stock
he wants to purchase by $S_{i-1}$
each. At the end of the round $i$ {\it Stock Market} announces
the price $S_i$ of the stock, and {\it Trader} receives his gain or
suffers loss $M_i(S_i-S_{i-1})$ for round $i$. The total gain or loss
for the first $n$ rounds is equal to $\sum\limits_{i=1}^n M_i(S_i-S_{i-1})$.

We show that, using the well-calibrated forecasts, it is possible
to construct a universal strategy for algorithmic trading in the stock market
which performs asymptotically at least as well as any stationary trading
strategy presented by a continuous function $D$ from the object ${\bf x}_i$.
This universal trading strategy is of decision type: we
buy or sell only one share of the stock at each round.
The learning process is the most traditional one. At each step,
{\it Trader} makes a randomized prediction $\tilde p_i$ of
a future price $S_i$ of the stock and takes ``the best response'' to this
prediction.
He chooses a strategy to going long: $\tilde M_i=1$
if $\tilde p_i>\tilde S_{i-1}$, or to going short: $\tilde M_i=-1$, otherwise,
where $\tilde S_{i-1}$ is the randomized past price of the stock.
{\it Trader} uses some randomized
algorithm for computing the well-calibrated forecasts $\tilde p_i$.

Therefore, our universal strategy uses some internal randomization.

{\it Trader M} can buy or sell only one share of the stock.
Therefore, in order to compare the performance of the traders we
have to standardize the strategy of {\it Trader D}.
We use the norm $\|D\|_\infty=\sup\limits_{0\le x\le 1} |D(x)|$ and
a normalization factor $\|D\|_+=\max\{1,\|D\|_\infty\}$,
where $D$ is a continuous function.
Our main result,
Theorems~\ref{main-r1} and~\ref{main-r1a} (Section~\ref{sec-2}), and
Theorem~\ref{iniversal-consist-1} (Section~\ref{universal-1}),
says that this trading strategy $\tilde M_i$ performs asymptotically
at least as well as any stationary trading strategy presented by a
continuous function $D(x)$.
With probability one, the gain of this trading strategy
is asymptotically not less than the average gain of any stationary
trading strategy $D$ from one share of the stock:
\begin{eqnarray}                               	
\liminf\limits_{n\to\infty}\left(\frac{1}{n}\sum\limits_{i=1}^n \tilde M_i(S_i-S_{i-1})-
\|D\|^{-1}_+\frac{1}{n}\sum\limits_{i=1}^n
D({\bf x}_i)(S_i-S_{i-1})\right)\ge 0,
\label{compete-stat-1}
\end{eqnarray}
where ${\bf x}_i$ is a side information used by the stationary
trading strategy $D$ at step $i$.

Evidently, the requirement (\ref{compete-stat-1})
for all continuous $D$ is equivalent to
the requirement:
\begin{eqnarray*}                               	
\liminf\limits_{n\to\infty}\left(\frac{1}{n}\sum\limits_{i=1}^n \tilde M_i(S_i-S_{i-1})-
\frac{1}{n}\sum\limits_{i=1}^n D({\bf x}_i)(S_i-S_{i-1})\right)\ge 0
\end{eqnarray*}
for all continuous $D$ such that $\|D\|_\infty\le 1$.

To achieve this goal we extend in Theorem~\ref{univ-1b}
(Section~\ref{gam-1a}) Kakade and Foster's forecasting algorithm
for a case of arbitrary real valued outcomes and to a more general notion of calibration
with changing parameterized checking rules. We combine
it with~\citet{TaV2005} defensive forecasting method in RKHS
(see~\citealt{Vov2005a}).
In Section~\ref{universal-1},
using a universal kernel, we generalize
this result to any continuous stationary trading strategy.
We show in Section~\ref{universal-2} that the universality
property fails if we consider discontinuous trading strategies.
On the other hand, we show in Theorem~\ref{main-r1des} that a universal
trading strategy exists for a class of randomized discontinuous trading
strategies.

In Section~\ref{app-1} results of numerical experiments are
presented. Our empirical results on historical markets provide
strong evidence that this type of algorithmic trading can "beat the
market": our universal strategy is always better than
``buy-and-hold'' strategy for each stock chosen arbitrarily in
Stock Market. This strategy outperforms also an algorithmic trading
strategy using some standard prediction algorithm (ARMA).

Some parts of this work were presented in \citet{Vyu2013} and \citet{VyT2013}.

\section{Preliminaries}\label{prem-1}

By a kernel function on a set $X$ we mean any function $K(x,y)$
which can be represented as a dot product $K(x,y)=(\Phi(x)\cdot\Phi(y))$,
where $\Phi$ is a mapping from $X$ to some Hilbert feature space.

The reproducing kernels are of special interest.
A Hilbert space $\cal F$ of real-valued functions on a compact
metric space $X$ is called RKHS (Reproducing Kernel Hilbert Space) on $X$ if
the evaluation functional $f\to f(x)$ is continuous for each $x\in X$.
Let $\|\cdot\|_{{\cal F}}$ be a norm in $\cal F$ and
$c_{{\cal F}}(x)=\sup\limits_{\|f\|_{{\cal F}}\le 1}|f(x)|$.
The embedding constant of $\cal F$ is defined
$c_{{\cal F}}=\sup\limits_x c_{{\cal F}}(x)$.
We consider RKHS $\cal F$ with
$c_{{\cal F}}<\infty$.

Let $X=[0,1]^m$ for $m\ge 1$.
An example of RKHS is the Sobolev space ${\cal F}=H^1([0,1])$, which
consists of absolutely continuous functions
$f:[0,1]\to{\cal R}$ with $\|f\|_{{\cal F}}<\infty$, where
$
\|f\|_{{\cal F}}=\sqrt{\int_0^1(f(t))^2dt+\int_0^1(f'(t))^2dt}.
$
For this space, $c_{{\cal F}}=\sqrt{\coth 1}$ (see~\citealt{Vov2005a}).

Let $\cal F$ be an RKHS on $X$ with the dot product $(f\cdot g)$
for $f,g\in\cal F$. By Riesz--Fisher theorem, for each $x\in X$
there exists $k_x\in\cal F$ such that $f(x)=(k_x\cdot f)$.

The reproducing kernel is defined $K(x,y)=(k_x\cdot k_y)$.
The main properties of the kernel: 1) $K(x,y)=K(y,x)$ for all $x,y\in X$
(symmetry property); 2) $\sum\limits_{i,j=1}^k\alpha_i\alpha_jK(x_i,x_j)\ge 0$
for all $k$, for all $x_i\in X$, and for all real numbers
$\alpha_i$, where $i=1,\dots,k$ (positive semidefinite property).

Conversely, a kernel defines RKHS: any symmetric, positive
semidefinite kernel function $K(x,y)$ defines some canonical RKHS
$\cal F$ and a mapping $\Phi:X\to\cal F$ such that
$K(x,y)=(\Phi(x)\cdot\Phi(y))$. Also, $c_{{\cal
F}}(x)=\|k_x\|_{{\cal F}}=\|\Phi(x)\|_{{\cal F}}$. The mapping
$\Phi(x)$ is also called ``feature map'' (see~\citealt{CST2000}, Chapter 3).

A function $f:X\to\cal R$ is induced by a kernel $K(x,y)$ if
there exists an element $g\in\cal F$ such that
$f(x)=(g\cdot\Phi(x))$. This definition is independent of a map
$\Phi$. For any continuous kernel $K(x,y)$, every induced
function $f$ is continuous (see~\citet{Ste2001}).
\footnote
{
It is Lipschitz continuous (with respect to some
semimetrics induced by the feature map
(\citealt{Ste2001}, Lemma 3).
}
In what follows we consider continuous kernels.
Therefore, all functions from canonical RKHS $\cal F$ are continuous.

For Sobolev space $H^1([0,1])$, the reproducing kernel is
$$
K(t,t')=(\cosh\min(t,t')\cosh\min(1-t,1-t'))/\sinh 1
$$
(see~\citealt{Vov2005a}).

Well known examples of kernels on $X=[0,1]^m$: Gaussian kernel
$K(\bar x,\bar y)=\exp\{-\frac{\|\bar x-\bar y\|^2}{\sigma^2}\}$,
where $\|\cdot\|$ is the Euclidian norm; $K(t,t')=\cos(\frac{\pi}{2}(t-t'))$,
where $m=1$ and $t,t'\in [0,1]$.

Other examples and details of
the kernel theory see in~\citet{SmS2002}.

Some special kernel corresponds to the method of randomization
defined below.
A random variable $\tilde y$ is called randomization of a real number
$y\in [0,1]$ if $E(\tilde y)=y$, where $E$ is the symbol of
mathematical expectation with respect to the corresponding to $\tilde y$
probability distribution.

We use a specific method of randomization of real numbers from
unit interval proposed by~\citet{KaF2004}.
Given positive integer number $K$ divide the interval $[0,1]$ on
subintervals of length $\Delta=1/K$
with rational endpoints $v_i=i\Delta$, where $i=0,1,\dots , K$.
Let $V$ denotes the set of these points.
Any number $p\in [0,1]$ can be represented as a linear
combination of two neighboring endpoints of $V$ defining
subinterval containing $p$~:
\begin{eqnarray}
p=\sum\limits_{v\in V}w_v(p)v=
w_{v_{i-1}}(p)v_{i-1}+w_{v_i}(p)v_i,
\label{round-1q}
\end{eqnarray}
where $p\in [v_{i-1},v_i]$, $i=\lfloor p^1/\Delta+1\rfloor$,
$w_{v_{i-1}}(p)=1-(p-v_{i-1})/\Delta$, and
$w_{v_i}(p)=1-(v_i-p)/\Delta$.
Define $w_v(p)=0$ for all other $v\in V$.
Define a random variable
\[
\tilde p=
  \left\{
    \begin{array}{l}
      v_{i-1} \mbox{ with probability } w_{v_{i-1}}(p)
    \\
      v_i \mbox{ with probability } w_{v_i}(p)
    \end{array}
  \right.
\]
Let $\bar w(p)=(w_v(p):v\in V)$ be a vector of probabilities of rounding.

For any $k$-dimensional vector ${\bar x}=(x_1,\dots ,x_k)\in [0,1]^k$,
we round each coordinate $x_s$, $s=1,\dots k$ to
$v_{j_s-1}$ with probability $w_{v_{j_s-1}}(x_s)$ and to $v_{j_s}$
with probability $w_{v_{j_s}}(x_s)$,
where $x_s\in [v_{j_s-1},v_{j_s}]$. Let $\tilde x$ be the corresponding
random vector.

Let $v=(v^1,\dots,~v^k)\in V^k$ and
$W_v(\bar x)=\prod_{s=1}^k w_{v^s}(x_s)$. For any $\bar x$,
let $\bar W(\bar x)=(W_v(\bar x):v\in V^k)$ be a vector of
probability distribution in $V^k$:
$\sum\limits_{v\in V^k}W_v(\bar x)=1$.
For $\bar x,\bar y\in [0,1]^k$, the dot product
$
K_1(\bar x,\bar x')=(\bar W(\bar x)\cdot\bar W(\bar x'))
$
is the symmetric positive semidefinite kernel function.

\section{Well-calibrated forecasting with side information}\label{gam-1a}

A universal trading strategy, which will be defined in Section~\ref{sec-2}, is
based on the well-calibrated forecasts of stock prices.
In this section we present a randomized algorithm for computing
well-calibrated forecasts using a side information.

A standard way to present any forecasting process is the game-theoretic
protocol. The basic online
prediction protocol has two players {\it Reality} and {\it Predictor}
(see Fig~\ref{fig-1}).

\begin{figure}
\fbox{%
\parbox{14.8cm}{%
{\bf Basic prediction protocol}.

\noindent{\bf FOR} $i=1,2\dots$
\\
{\it Reality} announces a signal ${\bf x}_i$.
\\
{\it Predictor} announces a forecast $p_i$.
\\
{\it Reality} announces an outcome $y_i\in [0,1]$.
\\
{\bf ENDFOR}
}%
}
\caption{Basic prediction protocol}\label{fig-1}
\end{figure}

At the beginning of each step $i$, {\it Predictor} is given some
data ${\bf x}_i$ relevant to predicting the following outcome $y_i$.
We call ${\bf x}_i$ a signal or a side information.
Signals are taken from the {\it object} space.

The outcomes $y_i$ are taken from an outcome space and predictions $p_i$
are taken from a prediction space.
In this paper an outcome is a real number from the unit interval $[0,1]$ and a
forecast is a single number from this interval (which can be
output of a random variable). We could interpret the forecast
$p_i$ as the mean value of a future outcome $y_i$ under some
unknown to us probability distribution in $[0,1]$.

{\it Reality} is called {\it oblivious} if an infinite sequence
of outcomes and signals $y_1,{\bf x}_1, y_2,{\bf x}_2,\dots$
is defined before the game starts and {\it Reality}
only reveals their next value $y_i,{\bf x}_i$ at each step $i$. In
this case the outcomes and signals do not depend on past predictions.
In case of {\it non oblivious} {\it Reality} this sequence is not
fixed in advance and any next value $y_i,{\bf x}_i$ can be output
of some measurable function from previous moves of {\it Predictor}, ie,
from past predictions $p_1,\dots ,p_{i-1}$.

In what follows we compare two types of forecasting algorithms:
randomized algorithms which we will construct and stationary
forecasting strategies which are continuous functions $D$ from
some RKHS using a side information as input.
We consider two type of predictors: $C$ and $D$, playing according
to the basic prediction protocol presented at Fig~\ref{fig-1}.

This protocol is {\it perfect-information}
for {\it Predictor C}. This means that {\it Predictor C} can
use other players moves so far. Past outcomes
and predictions are also known to {\it Reality} in
the perfect-information protocol.

{\it Predictor D} can use only a signal ${\bf x}_i$ that is given at
the beginning of any step $i$.
{\it Predictor D} uses a stationary prediction strategy $D({\bf x}_i)$,
where $D$ is a function whose input is the signal ${\bf x}_i$ and
output is the number of shares. We suppose that
${\bf x}_i$ is a real number from the unit interval. The number ${\bf x}_i$
can encode any information.
For example, it can be past outcomes and signals and even the future
outcome $y_i$.

{\it Predictor C} uses a randomized strategy which we will
define below.
We collect all information used for the internal
randomization in a vector $\bar x_i$. This vector can contain
any information known before the move of {\it Predictor C} at
step $i$: the signal ${\bf x}_i$, past outcomes and so on.

For example, in Section~\ref{sec-2}, the information is one-dimensional
vector $\bar x_i=y_{i-1}$ that is the past outcome,
in Section~\ref{universal-2}, $\bar x_i=(y_{i-1},{\bf x}_i)$ is the pair
of the past outcome and the signal.

In general, we suppose that
$\bar x_i$ is a vector of dimension $k\ge 1$: $\bar x_i\in [0,1]^k$.
We call it {\it an information} vector and assume that some method
for computing information vectors given past outcomes and signals is fixed.

We use the tests of calibration to measure the discrepancy
between predictions and outcomes. These tests use
{\it the checking rules}. We consider checking rules of more general type
than that used in the literature on asymptotic calibration.

For any subset $R\subseteq [0,1]^{k+1}$, define
the checking rule that is an indicator function:
\[
I_R(p,\bar x)=
  \left\{
    \begin{array}{l}
      1 \mbox{ if } (p,\bar x)\in R,
    \\
      0 \mbox{ otherwise, }
    \end{array}
  \right.
\]
where $\bar x$ is an $k$-dimensional vector.

In Section~\ref{gam-1a} we set $k=1$ and $R=\{(p,y):p>y\}$ or
$R=\{(p,y):p\le y\}$, where $p,y\in [0,1]$.
In Section~\ref{universal-2},
$k=2$ and a set $R$ is defined in a more complex way.

In the online prediction protocol defined on Fig~\ref{fig-1},
given $\Delta>0$, a sequence of forecasts $p_1,p_2,\dots$ is called
{\it $\Delta$-calibrated} for a sequences of outcomes $y_1,y_2,\dots$
and information vectors $\bar x_1,\bar x_2,\dots$ if
for any subset $R\subseteq [0,1]^{k+1}$ the following asymptotic inequality
holds:
\begin{eqnarray*}
\limsup\limits_{n\to\infty}\left|\frac{1}{n}\sum\limits_{i=1}^n
I_R(p_i,\bar x_i)(y_i-p_i)\right|\le\Delta.
\end{eqnarray*}
The sequence of forecasts is called {\it well-calibrated} if
\begin{eqnarray}
\lim\limits_{n\to\infty}\frac{1}{n}\sum\limits_{i=1}^n
I_R(p_i,\bar x_i)(y_i-p_i)=0.
\label{well-c-1}
\end{eqnarray}
If {\it Reality} is non oblivious and acts ``adversatively'', then,
as shown by~\citet{Oak85} and~\citet{Daw85}, any deterministic
forecasting algorithm will not always be calibrated. In case where $k=0$,
{\it Reality} can define their outcomes by the rule:
\[
y_i=
  \left\{
    \begin{array}{l}
      0 \mbox{ if } p_i>\frac{1}{2}
    \\
      1 \mbox{ otherwise.}
    \end{array}
  \right.
\]
Then any sequence of forecasts $p_1,p_2,\dots$ will not be calibrated
for the sequence of such outcomes $y_1,y_2,\dots$.
It is easy to verify that the condition (\ref{well-c-1}) fails for
$R=[0,\frac{1}{2}]$ or for $R=[\frac{1}{2},1]$.

Following the method of~\citet{FoV98}, at each step $i$,
using the past outcomes $y_1,\dots ,y_{i-1}$,
we will define a deterministic forecast $p_i$ and randomize
it to a random variable $\tilde p_i$ using the method
of randomization defined in Section~\ref{prem-1}. We also randomize
the information vector $\bar x_i$ to a random vector $\tilde x_i$.
We call this {\it sequential randomization}.

This sequential randomization generates for any $i$ a probability distribution
$Pr_i$ on the set of all finite sequences $p_1,\bar x_1,\dots ,p_i,\bar x_i$
of forecasts and information vectors.
In case of oblivious {\it Reality} this is simply the product distribution
which in their turn generates the overall probability distribution $Pr$
on the set of all infinite trajectories
$p_1,\bar x_1,p_2,\bar x_2,\dots$.
In case of non oblivious {\it Reality}, at any step $i$, a probability
distribution $Pr_i$ on $[0,1]^i$ exists such that the corresponding
method of randomization of $p_i$
is defined as conditional distribution $Pr_i(\cdot|p_1,\dots ,p_{i-1})$
on $[0,1]$. The overall probability distribution $Pr$ on the set
of all infinite trajectories generating these
$Pr_i$ can be defined by Ionescu--Tulcea theorem (see~\cite{Shi80}).

The following theorem on calibration with a side information
is the main tool for an analysis
presented in Sections~\ref{sec-2} and~\ref{universal-2}.
We will show that for any subset $R\subseteq [0,1]^{k+1}$,
with $Pr$-probability 1, the equality (\ref{well-c-1}) is valid,
where $p_i$ and $\bar x_i$ are replaced on their
randomized variants $\tilde p_i$ and $\tilde x_i$.

In the prediction protocol defined on
Fig~\ref{fig-1}, let $y_1,y_2,\dots$ be a sequence of outcomes and
${\bf x}_1,{\bf x}_2,\dots$ be the corresponding sequences of signals
given online.
We assume that a sequence of the information vectors
$\bar x_1,\bar x_2,\dots\in {\cal R}^k$ also be defined online.

Let also, $\cal F$ be an RKHS on $[0,1]$ with a kernel
$K_2({\bf x},{\bf x}')$ and a finite embedding constant $c_{{\cal F}}$.

\begin{theorem}\label{univ-1b}
For any $\epsilon>0$, an algorithm for computing
forecasts $p_1,p_2,\dots$ and a sequential method of randomization
can be constructed such that the following three items hold:
\begin{itemize}
\item{}
For any $n$, $R\subseteq [0,1]^{k+1}$, and $\delta>0$,
with probability at least $1-\delta$,
\begin{eqnarray}
\left|\sum_{i=1}^{n}
I_R(\tilde p_i,\tilde x_i)(y_i-\tilde p_i)\right|\le
22\left(\frac{k+1}{4}\right)^{\frac{2}{k+3}}
(c^2_{{\cal F}}+1)^{\frac{1}{k+3}}n^{1-\frac{1}{k+3}+\epsilon}+
\nonumber
\\
+\sqrt{\frac{n}{2}\ln\frac{2}{\delta}},
\label{call-1b}
\end{eqnarray}
where $\tilde p_1,\tilde p_2,\dots$ are the
corresponding randomizations of $p_1,p_2,\dots$ and
$\tilde x_1,\tilde x_2,\dots$ are the corresponding randomizations
of $k$-dimensional information vectors $\bar x_1,\bar x_2,\dots$;
\item{} For any $D\in\cal F$ and $n$,
\begin{eqnarray}
\left|\sum_{i=1}^{n}D({\bf x}_i)(y_i-p_i)\right|\le
\|D\|_{{\cal F}}\sqrt{(c^2_{{\cal F}}+1)n},
\label{call-1ab}
\end{eqnarray}
where ${\bf x}_1,{\bf x}_2,\dots$ are signals.
\item{}
For any $R\subseteq [0,1]^{k+1}$, with probability 1,
\begin{eqnarray}
\lim\limits_{n\to\infty}\frac{1}{n}\sum_{i=1}^{n}
I_R(\tilde p_i,\tilde x_i)(y_i-\tilde p_i)=0.
\label{call-1bi}
\end{eqnarray}
\end{itemize}
\end{theorem}
{\it Proof.}
At first, in Proposition~\ref{univ-1c} (below),
given $\Delta>0$, we modify a randomized rounding algorithm of
~\citet{KaF2004} to construct some $\Delta$-calibrated
forecasting algorithm, and combine it with~\citet{Vov2005a}
defensive forecasting algorithm. After that, we revise it
tending $\Delta\to 0$ such that (\ref{call-1b}) will hold.

\begin{proposition}\label{univ-1c}
Under the assumptions of Theorem~\ref{univ-1b},
an algorithm for computing forecasts and a method of randomization
can be constructed such that
the inequality (\ref{call-1ab}) holds for all $D$ from RKHS $\cal F$
and for all $n$. Also, for any $n$, $R$, and $\delta>0$,
with probability at least $1-\delta$,
\begin{eqnarray*}
\left|\sum_{i=1}^{n}
I_R(\tilde p_i,\tilde x_i)(y_i-\tilde p_i)\right|\le\Delta n+
\sqrt{\frac{n(c^2_{{\cal F}}+1)}{\Delta^k}}
+\sqrt{\frac{n}{2}\ln\frac{2}{\delta}}.
\end{eqnarray*}
\end{proposition}
{\it Proof}. We define a deterministic forecast and after that we randomize it.

The partition $V=\{v_0,\dots,v_K\}$ and probabilities of rounding
were defined above by (\ref{round-1q}).
In what follows we round some deterministic forecast $p_n$ to
$v_{i-1}$ with probability $w_{v_{i-1}}(p_n)$ and to $v_i$
with probability $w_{v_i}(p_n)$.
We also round each coordinate $x_{n,s}$, $s=1,\dots k$, of the
information vector $\bar x_n$ to
$v_{j_s-1}$ with probability $w_{v_{j_s-1}}(x_{n,s})$ and to $v_{j_s}$
with probability $w_{v_{j_s}}(x_{n,s})$,
where $x_{n,s}\in [v_{j_s-1},v_{j_s}]$.

Let $W_v(p_n,\bar x_n)=w_{v^1}(p_n)w_{v^2}(\bar x_n)$,
where $v=(v^1,v^2)$ and $v^1\in V$, $v^2=(v^2_1,\dots v^2_k)\in V^k$,
$w_{v^2}(\bar x_n)=\prod_{s=1}^k w_{v^2_s}(x_{n,s})$,
and $\bar W(p_n,\bar x_n)=(W_v(p_n,\bar x_n):v\in V^{k+1})$ be a vector of
probability distribution in $V^{k+1}$. Define the corresponding kernel
$K_1(p,\bar x,p',\bar x')=(\bar W(p,\bar x)\cdot \bar W(p',\bar x'))$.

Let the deterministic forecasts $p_1,\dots , p_{n-1}$ be already defined
(put $p_1=1/2$). We want to define a deterministic forecast $p_n$.

The kernel $K_2({\bf x},{\bf x}')$ can be represented
as a dot product in some feature space:
$K_2({\bf x},{\bf x}')=(\Phi({\bf x})\cdot\Phi({\bf x}')$.
Consider
\begin{eqnarray}
U_n(p)=\sum\limits_{i=1}^{n-1}(K_1(p,\bar x_n,p_i,\bar x_i)+
K_2({\bf x}_n,{\bf x}_i))(y_i-p_i).
\label{kernel-comb-1}
\end{eqnarray}

The following lemma presents a general method for computing the
deterministic forecasts.

Define ${\cal M}_0=1$ and
$$
{\cal M}_n={\cal M}_{n-1}+U_n(p_n)(y_n-p_n)
$$
for all $n$.

\begin{lemma}\label{lemmmm-1} (~\citealt{TaV2005})
A sequence of forecasts $p_1,p_2,\dots$ can be
computed such that ${\cal M}_n\le {\cal M}_{n-1}$ for all $n$.
\end{lemma}
{\it Proof}. By definition the function $U_n(p)$ is
continuous in $p$. The needed forecast is computed as follows.
If $U_n(p)>0$ for all $p\in[0,1]$ then
define $p_n=1$; if $U_n(p)<0$ for all $p\in[0,1]$ then
define $p_n=0$. Otherwise, define $p_n$ to be a root of the equation
$U_n(p)=0$ (some root exists by the intermediate value theorem).
Evidently, $M_n\le M_{n-1}$ for all $n$. Lemma is proved. $\triangle$

Now we continue the proof of the proposition.

Let forecasts $p_1,p_2,\dots$ be computed by the method of
Lemma~\ref{lemmmm-1}. Then for any $N$,
\begin{eqnarray}
0\ge {\cal M}_N-{\cal M}_0=\sum\limits_{n=1}^N
U_n(p_n)(y_n-p_n)=
\nonumber
\\
=\sum\limits_{n=1}^N\sum\limits_{i=1}^{n-1}
(K_1(p_n,\bar x_n,p_i,\bar x_i)+K_2({\bf x}_n,{\bf x}_i))
(y_i-p_i)(y_n-p_n)=
\nonumber
\\
=\frac{1}{2}\sum\limits_{n=1}^N\sum\limits_{i=1}^N
K_1(p_n,\bar x_n,p_i,\bar x_i)(y_i-p_i)(y_n-p_n)-
\nonumber
\\
-\frac{1}{2}\sum\limits_{n=1}^N (K_1(p_n,\bar x_n,p_n,\bar x_n)(y_n-p_n))^2+
\nonumber
\\
+\frac{1}{2}\sum\limits_{n=1}^N\sum\limits_{i=1}^N
K_2({\bf x}_n,{\bf x}_i)(y_i-p_i)(y_n-p_n)-
\nonumber
\\
-\frac{1}{2}\sum\limits_{n=1}^N (K_2({\bf x}_n,{\bf x}_n)(y_n-p_n))^2=
\\
=\frac{1}{2}\left\|\sum\limits_{n=1}^N
\bar W(p_n,\bar x_n)(y_n-p_n)\right\|^2-
\frac{1}{2}\sum\limits_{n=1}^N \|\bar W(p_n,\bar x_n)\|^2(y_n-p_n)^2+
\label{uun-2ii}
\\
+\frac{1}{2}\left\|\sum\limits_{n=1}^N
\Phi({\bf x}_n)(y_n-p_n)\right\|_{{\cal F}}^2-
\frac{1}{2}\sum\limits_{n=1}^N \|\Phi({\bf x}_n)\|_{{\cal F}}^2(y_n-p_n)^2.
\label{uun-2i}
\end{eqnarray}
In (\ref{uun-2ii}), $\|\cdot\|$ is Euclidian norm, and in
(\ref{uun-2i}), $\|\cdot\|_{{\cal F}}$ is the norm in RKHS $\cal F$.

Since $(y_n-p_n)^2\le 1$ for all $n$ and
\begin{eqnarray*}
\|(\bar W(p_n,\bar x_n)\|^2=\sum\limits_{v\in V^{k+1}}(W_v(p_n,\bar x_n))^2\le
\sum\limits_{v\in V^{k+1}}W_v(p_n,\bar x_n)=1,
\label{uun-2iig}
\end{eqnarray*}
the subtracted sum of (\ref{uun-2ii}) is upper bounded by $N$.

Since $\|\Phi({\bf x}_n)\|_{{\cal F}}=c_{{\cal F}}(\bar x_n)$ and
$c_{{\cal F}}({\bf x})\le c_{{\cal F}}$ for all $\bf x$,
the subtracted sum of (\ref{uun-2i}) is upper bounded by $c^2_{{\cal F}}N$.
As a result we obtain
\begin{eqnarray}
\left\|\sum\limits_{n=1}^N
\bar W(p_n,\bar x_n)(y_n-p_n)\right\|\le\sqrt{(c^2_{{\cal F}}+1)N}
\label{uun-2iiis}
\\
\left\|\sum\limits_{n=1}^N
\Phi({\bf x}_n)(y_n-p_n)\right\|_{{\cal F}}\le\sqrt{(c^2_{{\cal F}}+1)N}
\label{uun-2iii}
\end{eqnarray}
for all $N$. Let us denote
$
\bar\mu_n=\sum\limits_{i=1}^n\bar W(p_i,\bar x_i)(y_i-p_i).
$
By (\ref{uun-2iiis}), $\|\bar\mu_n\|\le\sqrt{(c^2_{{\cal F}}+1)n}$
for all $n$.

Let $\bar\mu_n=(\mu_n(v):v\in V^{k+1})$. By definition for any  $v$,
\begin{eqnarray}
\mu_n(v)=\sum\limits_{i=1}^n W_{v}(p_i,\bar x_i)(y_i-p_i).
\label{mmm-1}
\end{eqnarray}
Insert the term $I(v)$ in the sum (\ref{mmm-1}), where $I$ is
the characteristic function of an arbitrary set
${\cal S}\subseteq [0,1]^{k+1}$,
sum by $v\in V^{k+1}$, and exchange the order of
summation. Using Cauchy--Schwarz inequality for vectors
$\bar I=(I(v):v\in V^{k+1})$, $\bar\mu_n=(\mu_n(v):v\in V^{k+1})$ and
Euclidian norm, we obtain
\begin{eqnarray}\label{inn-2a}
\left|\sum\limits_{i=1}^n\sum\limits_{v\in V^{k+1}}W_{v}(p_i,\bar x_i)
I(v)(y_i-p_i)\right|=
\nonumber
\\
=\left|\sum\limits_{v\in V^{k+1}}I(v)
\sum\limits_{i=1}^n W_{v}(p_i,\bar x_i)(y_i-p_i)\right|=
\nonumber
\\
=(\bar I\cdot\bar\mu_n)\le\|\bar I\|\cdot\|\bar\mu_n\|
\le\sqrt{|V^{k+1}|(c^2_{{\cal F}}+1)n}
\end{eqnarray}
for all $n$, where $|V^{k+1}|=(1+\frac{1}{\Delta})^{k+1}\le
\left(\frac{2}{\Delta}\right)^{k+1}$ is the cardinality of
the partition.

Let $\tilde p_i$ be a random variable taking values
$v\in V$ with probabilities $w_{v}(p_i)$ (only two of them are nonzero).
Recall that $\tilde x_i$ is a random variable taking
values $v\in V^k$ with probabilities $w_{v}(\bar x_i)$.

Let ${\cal S}\subseteq [0,1]^{k+1}$ and $I$ be its indicator function.
For any $i$, the mathematical expectation of a random variable
$I(\tilde p_i,\tilde x_i)(y_i-\tilde p_i)$ is equal to
\begin{eqnarray}\label{expe-1}
E(I(\tilde p_{i},\tilde x_i)(y_i-\tilde p_i))=
\sum\limits_{v\in V^{k+1}}W_{v}(p_i,\bar x_i)I(v)(y_i-v^1),
\end{eqnarray}
where $v=(v^1,v^2)$.
By Azuma--Hoeffding inequality (see (\ref{iii-1ss}) below), for any $n$ and
$\delta>0$, with $Pr$-probability $1-\delta$,
\begin{eqnarray}
\left|\sum\limits_{i=1}^n I(\tilde p_{i},\tilde x_i)(y_i-\tilde p_i)-
\sum\limits_{i=1}^n E(I(\tilde p_{i},\tilde x_i)
(y_i-\tilde p_i))\right|\le\sqrt{\frac{n}{2}\ln\frac{2}{\delta}}.
\label{iii-1}
\end{eqnarray}
By definition of the deterministic forecast
\begin{eqnarray*}
\left|\sum\limits_{v\in V^{k+1}}W_{v}(p_i,\bar x_i)I(v)(y_i-p_i)-
\sum\limits_{v\in V^{k+1}}W_{v}(p_i,\bar x_i)I(v)(y_i-v^1)\right|<
\Delta
\end{eqnarray*}
for all $i$, where $v=(v^1,v^2)$.
Summing (\ref{expe-1}) over $i=1,\dots ,n$ and using the inequality
(\ref{inn-2a}), we obtain
\begin{eqnarray}
\left|\sum\limits_{i=1}^n
E(I(\tilde p_{i},\tilde x_i)(y_i-\tilde p_i))\right|=
\nonumber
\\
=\left|\sum\limits_{i=1}^n\sum\limits_{v\in V^{k+1}}
W_v(p_i,\bar x_i)I(v)(y_i-v^1)\right|<
\nonumber
\\
<\Delta n+\sqrt{(c^2_{{\cal F}}+1)n/\Delta^{k+1}}
\label{iinn-2}
\end{eqnarray}
for all $n$.

By (\ref{iii-1}) and (\ref{iinn-2}), with $Pr$-probability $1-\delta$,
\begin{eqnarray}
\left|\sum\limits_{i=1}^n
I(\tilde p_{i}, \tilde x_i)(y_i-\tilde p_i)\right|\le
\Delta n+\sqrt{(c^2_{{\cal F}}+1)n/\Delta^{k+1}}+
\sqrt{\frac{n}{2}\ln\frac{2}{\delta}}.
\label{con-1}
\end{eqnarray}
By Cauchy--Schwarz inequality:
\begin{eqnarray*}
\left|\sum\limits_{n=1}^N D(\bar x_n)(y_n-p_n)\right|=
\left|\sum\limits_{n=1}^N (y_n-p_n)(D\cdot\Phi(\bar x_n))\right|=
\\
\left|\left(\sum\limits_{n=1}^N (y_n-p_n)\Phi(\bar x_n)\cdot D\right)\right|\le
\left\|\sum\limits_{n=1}^N (y_n-p_n)\Phi(\bar x_n)\right\|_{{\cal F}}
\cdot \|D\|_{{\cal F}}\le
\\
\le\|D\|_{{\cal F}}\sqrt{(c^2_{{\cal F}}+1)N}.
\end{eqnarray*}
Proposition is proved. $\triangle$

Now we turn to the proof of Theorem~\ref{univ-1b}.

The expression
$\Delta n+\sqrt{(c^2_{{\cal F}}+1)n\left(\frac{2}{\Delta}\right)^{k+1}}$
from (\ref{iinn-2}) and (\ref{con-1}) takes its minimal value at
$\Delta=2(\frac{k+1}{4})^\frac{2}{k+3}(c^2_{{\cal F}}+1)^{\frac{1}{k+3}}
n^{-\frac{1}{k+3}}$.
In this case, the right-hand side of the inequality (\ref{iinn-2}) is equal to
\begin{eqnarray}
\Delta n+\sqrt{n(c^2_{{\cal F}}+1)\left(\frac{2}{\Delta}\right)^{k+1}}\le
2\Delta n=
4\left(\frac{k+1}{4}\right)^\frac{2}{k+3}(c^2_{{\cal F}}+1)^{\frac{1}{k+3}}
n^{1-\frac{1}{k+3}}.
\label{conm-1}
\end{eqnarray}
In what follows we use the upper bound $2\Delta n$ in (\ref{iinn-2}).

To prove the bound (\ref{call-1b}) choose a monotonic sequence of
rational numbers
$
\Delta_1>\Delta_2>\dots
$
such that $\Delta_s\to 0$ as $s\to\infty$.
We also define an increasing sequence of positive integer numbers
$
n_{1}<n_{2}<\dots
$
For any $s$, we use for randomization on steps $n_{s}\le n<n_{s+1}$
the partition of $[0,1]$ on subintervals of length $\Delta_{s}$.

We start our sequences from $n_1=1$ and $\Delta_1=1$.
Also, define the numbers $n_2,n_3,\dots$ such that the inequality
\begin{eqnarray}
\left|\sum\limits_{i=1}^{n} E(I(\tilde p_i,\tilde x_i)
(y_i-\tilde p_i))\right|\le 4(s+1)\Delta_{s} n
\label{exp1-1}
\end{eqnarray}
holds for all $n_s\le n\le n_{s+1}$ and for all $s\ge 1$.

We define this sequence by mathematical induction on $s$.
Suppose that $n_s$ ($s\ge 1$) is defined such that the inequality
\begin{eqnarray}
\left|\sum\limits_{i=1}^{n} E(I(\tilde p_i,\tilde x_i)
(y_i-\tilde p_i))\right|\le 4s\Delta_{s-1}n
\label{exp1-1g}
\end{eqnarray}
holds for all $n_{s-1}\le n\le n_s$, and the inequality
\begin{eqnarray}
\left|\sum\limits_{i=1}^{n_s} E(I(\tilde p_i,\tilde x_i)
(y_i-\tilde p_i))\right|\le 4s\Delta_{s} n_s
\label{exp1-gg1}
\end{eqnarray}
also holds.

Let us define $n_{s+1}$. Consider all forecasts $\tilde p_i$
defined by the algorithm given above for the discretization
$\Delta=\Delta_{s+1}$. We do not use first $n_s$ of these
forecasts (more correctly we will use them only in bounds
(\ref{exp1-1f}) and (\ref{exp1-1fg}); denote these forecasts
${\bf \hat p_1,\dots ,\hat p_{n_s}}$). We add the forecasts
$\tilde p_i$ for $i>n_s$ to the forecasts defined before this
step of induction (for $n_s$). Let $n_{s+1}$ be such that the
inequality
\begin{eqnarray}
\left|\sum\limits_{i=1}^{n_{s+1}} E(I(\tilde p_i,\tilde x_i)
(y_i-\tilde p_i))\right|\le
\left|\sum\limits_{i=1}^{n_s} E(I(\tilde p_i,\tilde x_i)
(y_i-\tilde p_i))\right|+
\nonumber
\\
+\left|\sum\limits_{i=n_s+1}^{n_{s+1}} E(I(\tilde p_i,\tilde x_i)
(y_i-\tilde p_i))+
\sum\limits_{i=1}^{n_s} E(I({\bf \hat p_i},\tilde x_i)
(y_i-{\bf \hat p_i}))\right|+
\nonumber
\\
+\left|\sum\limits_{i=1}^{n_s} E(I({\bf\hat p_i},\tilde x_i)
(y_i-{\bf \hat p_i}))\right|
\le 4(s+1)\Delta_{s+1}n_{s+1}~~~~~
\label{exp1-1f}
\end{eqnarray}
holds. Here the first sum of the right-hand side of the
inequality (\ref{exp1-1f}) is bounded by $4s\Delta_{s}n_{s}$
-- by the induction hypothesis (\ref{exp1-gg1}). The second and
third sums are bounded by $2\Delta_{s+1}n_{s+1}$ and by
$2\Delta_{s+1}n_s$, respectively, where $\Delta=\Delta_{s+1}$ is defined
such that (\ref{conm-1}) holds. This follows from (\ref{iinn-2})
and by choice of $n_s$.

The induction hypothesis (\ref{exp1-gg1}) is valid for
$$
n_{s+1}\ge\frac{2s\Delta_s+\Delta_{s+1}}{\Delta_{s+1}(2s+1)}n_s.
$$
Similarly,
\begin{eqnarray}
\left|\sum\limits_{i=1}^{n} E(I(\tilde p_i,\tilde x_i)
(y_i-\tilde p_i))\right|\le
\left|\sum\limits_{i=1}^{n_s} E(I(\tilde p_i,\tilde x_i)
(y_i-\tilde p_i))\right|+
\nonumber
\\
+\left|\sum\limits_{i=n_s+1}^{n} E(I(\tilde p_i,\tilde x_i)
(y_i-\tilde p_i))+
\sum\limits_{i=1}^{n_s} E(I({\bf \hat p_i},\tilde x_i)
(y_i-{\bf \hat p_i}))\right|+
\nonumber
\\
+\left|\sum\limits_{i=1}^{n_s} E(I(\hat p_i,\tilde x_i)
(y_i-{\bf\hat p_i}))\right|
\le 4(s+1)\Delta_sn~~~~~
\label{exp1-1fg}
\end{eqnarray}
for $n_s<n\le n_{s+1}$. Here the first sum of the right-hand
inequality (\ref{exp1-1f}) is also bounded by
$4s\Delta_{s}n_{s}\le 4s\Delta_{s}n$ -- by the induction
hypothesis (\ref{exp1-gg1}). The second and the third sums are
bounded by $2\Delta_{s+1}n\le 2\Delta_sn$ and by
$2\Delta_{s+1}n_s\le 2\Delta_sn$, respectively. This follows
from (\ref{iinn-2}) and from choice of $\Delta_s$. The
induction hypothesis (\ref{exp1-1g}) is valid.

By (\ref{exp1-1}) for any $s$
\begin{eqnarray}
\left|\sum\limits_{i=1}^{n} E(I(\tilde p_i,\tilde x_i)
(y_i-\tilde p_i))\right|\le 4(s+1)\Delta_sn
\label{exp1-1er}
\end{eqnarray}
for all $n\ge n_s$ if $\Delta_{s}$ satisfies
the condition $\Delta_{s+1}\le\Delta_{s}(1-\frac{1}{s+2})$ for all $s$.

We show now that sequences $n_{s}$ and $\Delta_{s}$ satisfying
all the conditions above exist.

Let $\epsilon>0$ and $M=\lceil 2/\epsilon\rceil$, where
$\lceil r\rceil$ is the least integer number such that $m\ge r$.
Define $n_s=(s+M)^M$ and
$
\Delta_s=2\left(\frac{k+1}{4}\right)^\frac{2}{k+3}
(c^2_{{\cal F}}+1)^{\frac{1}{k+3}}n_s^{-\frac{1}{k+3}}.
$
Easy to verify that all requirements for $n_s$ and $\Delta_s$
given above are satisfied for all $s\ge s_0$, where $s_0$ is
sufficiently large. We redefine $n_i=n_{s_0}$ for all $1\le i\le s_0$.
Then all these requirements hold for these $i$ trivially.

We have in (\ref{exp1-1er}) for all $n_s\le n<n_{s+1}$
\begin{eqnarray*}
4(s+1)\Delta_s n\le 4(s+M)\Delta_s n_{s+1}=
\\
=8\left(\frac{k+1}{4}\right)^{\frac{2}{k+3}}(c^2_{{\cal F}}+1)^{\frac{1}{k+3}}
(s+M)(s+M+1)^M (s+M)^{-\frac{M}{k+3}}\le
\\
\le 22\left(\frac{k+1}{4}\right)^{\frac{2}{k+3}}
(c^2_{{\cal F}}+1)^{\frac{1}{k+3}}n_s^{1-\frac{1}{k+3}+2/M}\le
\\
\le 22\left(\frac{k+1}{4}\right)^{\frac{2}{k+3}}
(c^2_{{\cal F}}+1)^{\frac{1}{k+3}}n^{1-\frac{1}{k+3}+\epsilon}.
\end{eqnarray*}
Therefore, we obtain
\begin{eqnarray}
\left|\sum\limits_{i=1}^{n} E(I(\tilde p_i,\tilde x_i)
(y_i-\tilde p_i))\right|\le
22\left(\frac{k+1}{4}\right)^{\frac{2}{k+3}}
(c^2_{{\cal F}}+1)^{\frac{1}{k+3}}n^{1-\frac{1}{k+3}+\epsilon}
\label{exp1-1eri}
\end{eqnarray}
for all $n$.
Azuma--Hoeffding inequality says that for any $\gamma>0$
\begin{eqnarray}
Pr\left\{\left|\frac{1}{n}\sum\limits_{i=1}^n V_i\right|>\gamma
\right\}\le 2e^{-2n\gamma^2}
\label{iii-1ss}
\end{eqnarray}
for all $n$, where $V_i$ are martingale--differences.

We set $V_i=I(\tilde p_i,\tilde x_i)(y_i-\tilde p_i)-
E(I(\tilde p_i,\tilde x_i)(y_i-\tilde p_i))$ and
$\gamma=\sqrt{\frac{1}{2n}\ln\frac{2}{\delta}}$,
where $\delta>0$. Denote $\nu(n)=22\left(\frac{k+1}{4}\right)^{\frac{2}{k+3}}
(c^2_{{\cal F}}+1)^{\frac{1}{k+3}}n^{1-\frac{1}{k+3}+\epsilon}$.

Combining (\ref{exp1-1eri}) with (\ref{iii-1ss}), we obtain that
for any $n$ and $\delta>0$, with probability $1-\delta$,
\begin{eqnarray*}
\left|\sum\limits_{i=1}^n
I(\tilde p_i,\tilde x_i)(y_i-\tilde p_i)\right|\le
\nu(n)+\sqrt{\frac{n}{2}\ln\frac{2}{\delta}}.
\end{eqnarray*}

The asymptotic relation (\ref{call-1bi}) follows from (\ref{call-1b})
by Borel--Cantelli lemma. The proof is similar to the final part of the proof
of Theorem~\ref{main-r1a} below.
Theorem~\ref{univ-1b} is proved. $\triangle$

\section{Competing with stationary trading strategies from RKHS}\label{sec-2}

A trading game has two players: {\it Trader} and {\it Stock Market}.
They correspond to {\it Predictor} and {\it Reality} in the simple
prediction game defined in Section~\ref{gam-1a}.

We suppose that the prices $S_1,S_2,\dots$ of a stock are bounded and
rescaled such that $0\le S_i\le 1$ for all $t$. We set also $S_0=0$.
These prices are analogs of outcomes of the prediction game.

We present the process of algorithmic trading in Stock Market in the form of
a trading game regulated by the perfect-information protocol
presented on Fig~\ref{fig-3}.
\begin{figure}
\fbox{%
\parbox{14.8cm}{%
{\bf Basic trading protocol}.

\noindent{\bf FOR} $i=1,2\dots$
\\
{\it Stock Market} announces a signal ${\bf x}_i\in X$.
\\
{\it Trader} bets by buying or selling a number $C_i$ of shares of
the stock by $S_{i-1}$ each.
\\
{\it Stock Market} reveals a price $S_i$ of the stock.
\\
{\it Trader} receives his total gain (or suffers loss)
at the end of step $i$~:
\\
${\cal K}_i={\cal K}_{i-1}+C_i (S_i-S_{i-1})$.
We set ${\cal K}_0=0$.
\\
{\bf ENDFOR}
}%
}
\caption{Basic trading protocol}\label{fig-3}
\end{figure}

At the beginning of each step $i$ {\it Trader} is given an object
${\bf x}_i\in X$ which was called a side information at step $i$.
Without loss of generality suppose that $X=[0,1]$.

We call any sequence $\tilde M_i$, $i=1,2,\dots$, of random variables
{\it a randomized trading strategy}.
In case $\tilde M_i>0$ {\it Trader} playing for a rise, in case
$\tilde M_i<0$ {\it Trader} playing for a fall, {\it Trader} passes the step
if $C_i=0$.

We suppose that {\it Trader} buys $C_i$ shares (if $C_i>0$) or sells
$C_i$ shares (if $C_i\le 0$) at the beginning of any round $i$ and
sells or buys them at the end this round correspondingly.
Thus, {\it Trader} receives the gain or suffers the loss in the amount
of $C_i (S_i-S_{i-1})$ money units.

We suppose also that {\it Trader} can borrow money for buying shares
and can incur debt.

{\it A stationary trading strategy} is a function $D$ from $X$ to $\cal R$.
We suppose that some RKHS $\cal F$ on $X=[0,1]$
with a kernel $K_2({\bf x},{\bf x}')$ and with a finite embedding constant
$c_{{\cal F}}$ be given.

Any stationary trading strategy $D$ uses at step $i$ a side information
that is a real number ${\bf x}_i\in X$.

Our universal trading strategy will be randomized.
The universal trading strategy, which we define
below, uses the past price $S_{i-1}$ of the stock as one-dimensional
information vector in sense of Theorem~\ref{univ-1b}, where $S_0=0$.
This information is used for the internal randomization.

We define a universal trading strategy as a sequence of random variables
$\tilde M_i$ and show that
this trading strategy performs almost surely at least as well as any
stationary trading strategy $D\in\cal F$ using arbitrary side
information ${\bf x}_i$.

To be more concise, define on Fig~\ref{fig-4} the perfect-information
protocol of the game with two traders:
{\it Trader M} uses the randomized strategy $\tilde M_i$,
{\it Trader D} uses an arbitrary stationary trading strategy $D\in\cal F$.

\begin{figure}
\fbox{%
\parbox{14.8cm}{%
{\bf Trading protocol with two traders}.

\noindent{\bf FOR} $i=1,2\dots$
\\
{\it Stock Market} announces a signal ${\bf x}_i$.
\\
{\it Trader M} bets by buying or selling the random number $\tilde M_i$ of shares of
the stock by $S_{i-1}$ each.
\\
{\it Trader D} bets by buying or selling a number $D({\bf x}_i)$ of shares of
the stock by $S_{i-1}$ each.
\\
{\it Stock Market} reveals a price $S_i$ of the stock.
\\
{\it Trader M} receives his total gain (or suffers loss)
at the end of step $i$~:
\\
${\cal K}_i^M={\cal K}_{i-1}^M+\tilde M_i (S_i-S_{i-1})$.
We set ${\cal K}_0^M=0$.
\\
{\it Trader D} receives his total gain (or suffers loss)
at the end of step $i$~:
\\
${\cal K}_i^D={\cal K}_{i-1}^D+D({\bf x}_i)(S_i-S_{i-1})$.
We set ${\cal K}_0^D=0$.
\\
{\bf ENDFOR}
}%
}
\caption{Trading protocol with two traders}\label{fig-4}
\end{figure}

This protocol is more general than two basic trading protocols
(Fig~\ref{fig-3}) together, since {\it Stock Market}
can use information on the decisions of both traders $M$ and $D$
before revealing a future price $S_i$.

Past prices, signals and predictions are also known to {\it Trader M} in
the perfect-information protocol. {\it Trader D} can use only
side information. For example, at any step $i$,
past prices and predictions can be encoded in the signal ${\bf x}_i$
and used by {\it Trader D}.

At first, for simplicity, we consider a case of going long,
since the proof of optimality (Theorem~\ref{main-r1}) is much more clear
in this case than that in general case (Theorem~\ref{main-r1a}).
Also, a series of numerical experiments presented in Section~\ref{app-1},
are performed for the case where both traders going long.
The case of going short is considered similarly.

At each step $i$
we will compute a forecast $p_i$ of a future price and randomize it to
$\tilde p_i$. We also randomize the past price $S_{i-1}$ of the stock
to $\tilde S_{i-1}$. Details of this computation and randomization
are given in Section~\ref{gam-1a}. Our universal strategy is a randomized
decision rule -- it takes only two values:
\[
\tilde M_i^1=
\left\{
    \begin{array}{l}
      1 \mbox{ if } \tilde p_i>\tilde S_{i-1},
    \\
      0 \mbox{ otherwise. }
    \end{array}
  \right.
\]

Assume that prices $S_1,S_2,\dots\in [0,1]$ and signals
${\bf x}_1,{\bf x}_2,\dots\in [0,1]$
be given online according to the protocol presented on Fig~\ref{fig-4}.
Denote $\Delta S_i=S_i-S_{i-1}$.

Since {\it Trader M} can buy or sell only one share of the stock,
we have to standardize the strategy of {\it Trader D}.
We will use the norm
$\|D\|_\infty=\sup\limits_{{\bf x}\in [0,1]}|D({\bf x})|$ and
the normalization factor $\|D\|_+=\max\{1,\|D\|_\infty\}$
where $D$ is a nonnegative continuous function.

Informally, Theorem~\ref{main-r1} says that if the forecasts $\tilde p_i$
are well-calibrated for the sequence of prices $S_i$, $i=1,2,\dots$, then
{\it Trader M}, using the strategy $\tilde M_i^1$, performs
at least as well as any trader who going long using
a stationary trading strategy $D\in\cal F$.

\begin{theorem}\label{main-r1}
An algorithm for computing forecasts $p_i$ and a sequential
method of randomization can be constructed such that for any
nonnegative stationary trading strategy $D\in\cal F$
\begin{eqnarray}
\liminf\limits_{n\to\infty}\left(\frac{1}{n}\sum\limits_{i=1}^n
\tilde M_i^1
\Delta S_i-\frac{1}{n}\|D\|^{-1}_+
\sum\limits_{i=1}^n D({\bf x}_i)\Delta S_i\right)\ge 0
\label{asympt-1}
\end{eqnarray}
holds almost surely with respect to a probability distribution
generated by the corresponding sequential randomization.

Moreover, for any $\epsilon>0$ this trading strategy $\tilde M^1$
can be tuned such that
for any $n$ and $\delta>0$, with probability at least $1-\delta$, for
all nonnegative $D\in\cal F$,
\begin{eqnarray}                               	
\sum\limits_{i=1}^n \tilde M_i^1
\Delta S_i\ge \|D\|^{-1}_+\sum\limits_{i=1}^n
D({\bf x}_i)\Delta S_i-
\nonumber
\\
-30(c^2_{{\cal F}}+1)^{\frac{1}{4}}n^{\frac{3}{4}+\epsilon}-
\|D\|^{-1}_+\|D\|_{{\cal F}}\sqrt{(c^2_{{\cal F}}+1)n}-
\nonumber
\\
-\sqrt{\frac{n}{2}\ln\frac{2}{\delta}}.
\label{asympt-1a}
\end{eqnarray}
\end{theorem}
{\it Proof}. We use the randomized trading strategy $\tilde M^1$
based on the well-calibrated forecasts defined in Section~\ref{gam-1a},
where $y_i=S_i$ and $\bar x_i=S_{i-1}$.

Recall
that at any step $n$ we compute the deterministic forecast $p_n$
defined in Section~\ref{gam-1a} and its randomization to
$\tilde p_n$ using parameters $\Delta=\Delta_s=(c_{{\cal
F}}+1)^{\frac{1}{4}}(s+M)^{-\frac{M}{4}}$ and $n_s=(s+M)^M$,
where $n_s\le i<n_{s+1}$. Let also, $\tilde S_{i-1}$ be a
randomization of the past price $S_{i-1}$. The following upper
bound directly follows from the method of discretization:
\begin{eqnarray}
\left|\sum\limits_{i=1}^nI(\tilde p_i>\tilde S_{i-1})
(\tilde S_{i-1}-S_{i-1})\right|\le
\sum\limits_{t=0}^s(n_{t+1}-n_t)\Delta_t\le
\nonumber
\\
\le 4(c^2_{{\cal F}}+1)^{\frac{1}{4}}n_s^{\frac{3}{4}+\epsilon}\le
4(c^2_{{\cal F}}+1)^{\frac{1}{4}}n^{\frac{3}{4}}.
\label{esima-1}
\end{eqnarray}
Let $D({\bf x})$ be an arbitrary nonnegative trading strategy
from RKHS $\cal F$.
Clearly, the bound (\ref{esima-1}) holds if we replace
$I(\tilde p_i>\tilde S_{i-1})$
on $\|D\|^{-1}_+D({\bf x}_i)$.

Let $\tilde M^1$ be the randomized trading strategy
defined above. We use abbreviations:
\begin{eqnarray}
\nu_1(n)=4(c^2_{{\cal F}}+1)^{\frac{1}{4}}n^{\frac{3}{4}},
\label{hoeff-1}
\\
\nu_2(n)=
18n^{\frac{3}{4}+\epsilon}(c^2_{{\cal F}}+1)^{\frac{1}{4}}+
\sqrt{\frac{n}{2}\ln\frac{2}{\delta}},
\label{call-1bu}
\\
\nu_3(n)=\sqrt{(c^2_{{\cal F}}+1)n}
\label{call-1abu}
\end{eqnarray}
All sums below are for $i=1,\dots n$. By definition
$0\le D({\bf x}_i)\le\|D\|_+$ for all ${\bf x}_i\in [0,1]$.

Let $\delta>0$ and $n$ be given.
Then, with probability $1-\delta$, for any $D\in\cal F$,
the following chain of equalities and inequalities is valid:
\begin{eqnarray}
\sum\limits_{i=1}^n \tilde M^1_i
(S_i-S_{i-1})=
\sum\limits_{\tilde p_i>\tilde S_{i-1}}(S_i-S_{i-1})=
\nonumber
\\
=\sum\limits_{\tilde p_i>\tilde S_{i-1}}(S_i-\tilde p_i)+
\sum\limits_{\tilde p_i>\tilde S_{i-1}}(\tilde p_i-\tilde S_{i-1})+
\sum\limits_{\tilde p_i>\tilde S_{i-1}}(\tilde S_{i-1}-S_{i-1})\ge
\label{azu-1}
\\
\ge\sum\limits_{\tilde p_i>\tilde S_{i-1}}
(\tilde p_i-\tilde S_{i-1})-\nu_1(n)-\nu_2(n)\ge
\label{azu-2}
\\
\ge\|D\|^{-1}_+\sum\limits_{i=1}^n D({\bf x}_i)
(\tilde p_i-\tilde S_{i-1})-\nu_1(n)-\nu_2(n)=
\label{azu-2h}
\\
=\|D\|^{-1}_+\sum\limits_{i=1}^n D({\bf x}_i)(p_i-S_{i-1})-
\|D\|^{-1}_+\sum\limits_{i=1}^n D({\bf x}_i)(p_i-\tilde p_i)-
\nonumber
\\
-\|D\|^{-1}_+\sum\limits_{i=1}^n D({\bf x}_i)(\tilde S_{i-1}-S_{i-1})-
\nu_1(n)-\nu_2(n)\ge
\label{azu-3}
\\
\ge\|D\|^{-1}_+\sum\limits_{i=1}^n D({\bf x_i})(p_i-S_{i-1})-3\nu_1(n)-
\nu_2(n)=
\label{azu-4}
\\
=\|D\|^{-1}_+\sum\limits_{i=1}^n D({\bf x}_i)(S_i-S_{i-1})-
\|D\|^{-1}_+\sum\limits_{i=1}^n D({\bf x}_i)(S_i-p_i)-
\nonumber
\\
-3\nu_1(n)-\nu_2(n)-\|D\|^{-1}_+
\|D\|_{{\cal F}}\nu_3(n)\ge
\label{zzz-1}
\\
\ge\|D\|^{-1}_+\sum\limits_{i=1}^n D({\bf x}_i)(S_i-S_{i-1})-
\nonumber
\\
-3\nu_1(n)-\nu_2(n)-\|D\|^{-1}_+\|D\|_{{\cal F}}\nu_3(n).
\label{zzz-2}
\end{eqnarray}
In transition from (\ref{azu-1}) to (\ref{azu-2}) the inequality
(\ref{call-1b}) of Theorem~\ref{univ-1b} and the bound
(\ref{esima-1}) were used,
and so, the terms (\ref{hoeff-1}) and (\ref{call-1bu})
were subtracted. The transition from (\ref{azu-2}) to (\ref{azu-2h})
is valid since $0\le D({\bf x})\le\|D\|_+$ for all $\bf x$.
In transition from (\ref{azu-3}) to (\ref{azu-4})
the bound (\ref{esima-1})
was applied twice to intermediate terms, and so,
the term (\ref{esima-1})
was subtracted twice. In transition from (\ref{azu-4})
to (\ref{zzz-1}) the inequality (\ref{call-1ab}) of Theorem~\ref{univ-1b}
was used, and so, the term (\ref{call-1abu}) was subtracted.
In transition from (\ref{zzz-1}) to (\ref{zzz-2}) we have used
the inequality (\ref{call-1ab}) of Theorem~\ref{univ-1b}.
Therefore, we have (\ref{asympt-1a}).

The inequality (\ref{asympt-1}) follows from (\ref{asympt-1a})
by Borel--Cantelli lemma (see the final part of the proof
of Theorem~\ref{main-r1a} below). Theorem~\ref{main-r1} is proved. $\triangle$

Now, we consider the general case of going long and going short.
The corresponding universal trading strategy is defined:
\[
\tilde M_i=
\left\{
    \begin{array}{l}
      1 \mbox{ if } \tilde p_i>\tilde S_{i-1},
    \\
      -1 \mbox{ if } \tilde p_i\le\tilde S_{i-1}.
    \end{array}
  \right.
\]
{\it Trader $D$} is also can going long and short.

Let $S_1,S_2,\dots\in [0,1]$ and ${\bf x}_1,{\bf x}_2,\dots\in [0,1]$
be given online according to the protocol presented on Fig~\ref{fig-4}.

Informally, Theorem~\ref{main-r1a} says that if the forecasts $\tilde p_i$
are well-calibrated for the sequence of prices $S_i$, $i=1,2,\dots$, then
{\it Trader M}, using the strategy $\tilde M_i$, performs
at least as well as any trader who going long or short using
a stationary trading strategy $D\in\cal F$.

\begin{theorem}\label{main-r1a}
An algorithm for computing forecasts $p_i$ and a sequential method of
randomization can be constructed such that for any
stationary trading strategy $D\in\cal F$,
\begin{eqnarray}
\liminf\limits_{n\to\infty}\left(\frac{1}{n}\sum\limits_{i=1}^n
\tilde M_i
\Delta S_i-\frac{1}{n}\|D\|^{-1}_+
\sum\limits_{i=1}^n D({\bf x}_i)\Delta S_i\right)\ge 0
\label{asympt-1f}
\end{eqnarray}
holds almost surely with respect to a probability distribution
generated by the corresponding sequential randomization.

Moreover, for any $\epsilon>0$ this trading strategy $M$
can be tuned such that for any $n$ and $\delta>0$,
with probability at least $1-\delta$, for all $D\in\cal F$,
\begin{eqnarray}                               	
\sum\limits_{i=1}^n \tilde M_i
\Delta S_i\ge\|D\|^{-1}_+\sum\limits_{i=1}^n D({\bf x}_i)\Delta S_i-
\nonumber
\\
-52(c^2_{{\cal F}}+1)^{\frac{1}{4}}n^{\frac{3}{4}+
\epsilon}-\|D\|^{-1}_+\|D\|_{{\cal F}}\sqrt{(c^2_{{\cal F}}+1)n}-
\nonumber
\\
-2\sqrt{\frac{n}{2}
\ln\frac{2}{\delta}}.
\label{asympt-1abf}
\end{eqnarray}
\end{theorem}
{\it Proof}. We use abbreviations (\ref{hoeff-1})--(\ref{call-1abu})
from the proof of Theorem~\ref{main-r1}. Define
\[
D^+({\bf x})=
\left\{
    \begin{array}{l}
      D({\bf x}) \mbox{ if } D({\bf x})>0,
    \\
      0 \mbox{ otherwise. }
    \end{array}
  \right.
\]
and
\[
D^-({\bf x})=
\left\{
    \begin{array}{l}
      D({\bf x}) \mbox{ if } D({\bf x})\le 0,
    \\
      0 \mbox{ otherwise. }
    \end{array}
  \right.
\]
By definition $D({\bf x})=D^+({\bf x})+D^-({\bf x})$.

The proof of Theorem~\ref{main-r1a} is based on transformations
similar to (\ref{azu-1})--(\ref{zzz-2}).

Let $\delta>0$ and $n$ be given.
Then, with probability $1-\delta$, for any $D\in\cal F$,
\begin{eqnarray}
\sum\limits_{i=1}^n \tilde M_i
(S_i-S_{i-1})=
\nonumber
\\
=\sum\limits_{\tilde p_i>\tilde S_{i-1}}(S_i-S_{i-1})-
\sum\limits_{\tilde p_i\le\tilde S_{i-1}}(S_i-S_{i-1})=
\nonumber
\\
=\sum\limits_{\tilde p_i>\tilde S_{i-1}}(S_i-\tilde p_i)+
\sum\limits_{\tilde p_i>\tilde S_{i-1}}(\tilde p_i-\tilde S_{i-1})+
\sum\limits_{\tilde p_i>\tilde S_{i-1}}(\tilde S_{i-1}-S_{i-1})-
\nonumber
\\
-\sum\limits_{\tilde p_i\le\tilde S_{i-1}}(S_i-\tilde p_i)-
\sum\limits_{\tilde p_i\le\tilde S_{i-1}}(\tilde p_i-\tilde S_{i-1})-
\sum\limits_{\tilde p_i\le\tilde S_{i-1}}(\tilde S_{i-1}-S_{i-1})\ge
\nonumber
\\
\ge\sum\limits_{\tilde p_i>\tilde S_{i-1}}
(\tilde p_i-\tilde S_{i-1})-\nu_1(n)-\nu_2(n)-
\label{jjj-1}
\\
-\sum\limits_{\tilde p_i\le\tilde S_{i-1}}(\tilde p_i-\tilde S_{i-1})-
\nu_1(n)-\nu_2(n)\ge
\label{kkk-1}
\\
\ge\|D\|^{-1}_+\sum\limits_{\tilde p_i>\tilde S_{i-1}} D^+({\bf x}_i)
(\tilde p_i-\tilde S_{i-1})-
\nu_1(n)-\nu_2(n)+
\label{jjj-2}
\\
+\|D\|^{-1}_+\sum\limits_{\tilde p_i\le\tilde S_{i-1}} D^-({\bf x}_i)
(\tilde p_i-\tilde S_{i-1})-\nu_1(n)-\nu_2(n)=
\label{kkk-2}
\\
\ge\|D\|^{-1}_+\sum\limits_{i=1}^n D^+({\bf x}_i)
(\tilde p_i-\tilde S_{i-1})-
\nu_1(n)-\nu_2(n)+
\nonumber
\\
+\|D\|^{-1}_+\sum\limits_{i=1}^n D^-({\bf x}_i)
(\tilde p_i-\tilde S_{i-1})-\nu_1(n)-\nu_2(n)=
\nonumber
\\
=\|D\|^{-1}_+\sum\limits_{i=1}^n D({\bf x}_i)(\tilde p_i-\tilde S_{i-1})-
2\nu_1(n)-2\nu_2(n)=
\nonumber
\\
=\|D\|^{-1}_+\sum\limits_{i=1}^n D({\bf x}_i)(p_i-S_{i-1})-
\|D\|^{-1}_+\sum\limits_{i=1}^n D({\bf x}_i)(p_i-\tilde p_i)-
\nonumber
\\
-\|D\|^{-1}_+\sum\limits_{i=1}^n D({\bf x}_i)(\tilde S_{i-1}-S_{i-1})-
2\nu_1(n)-2\nu_2(n)\ge
\nonumber
\\
\ge\|D\|^{-1}_+\sum\limits_{i=1}^n D({\bf x_i})(p_i-S_{i-1})-
4\nu_1(n)-2\nu_2(n)=
\nonumber
\\
=\|D\|^{-1}_+\sum\limits_{i=1}^n D({\bf x}_i)(S_i-S_{i-1})-
\|D\|^{-1}_+\sum\limits_{i=1}^n D({\bf x}_i)(S_i-p_i)-
\nonumber
\\
-4\nu_1(n)-2\nu_2(n)-\|D\|^{-1}_+\|D\|_{{\cal F}}\nu_3(n)\ge
\nonumber
\\
\ge\|D\|^{-1}_+\sum\limits_{i=1}^n D({\bf x}_i)(S_i-S_{i-1})-
\nonumber
\\
-4\nu_1(n)-2\nu_2(n)-\|D\|^{-1}_+\|D\|_{{\cal F}}\nu_3(n).
\nonumber
\end{eqnarray}
The proof of these transitions is similar
to the proof of transitions in (\ref{azu-1})--(\ref{zzz-2}) of
Theorem~\ref{main-r1}.

To prove (\ref{asympt-1f}) we turn to Azuma--Hoeffding
inequality (\ref{iii-1ss}).
Denote $\gamma=\sqrt{\frac{1}{2n}\ln\frac{2}{\delta}}$.
Then $\delta=2e^{-n\gamma^2}$. Rewrite (\ref{asympt-1abf}) in the form:
\begin{eqnarray}                               	
\frac{1}{n}\sum\limits_{i=1}^n \tilde M_i^1
\Delta S_i-\frac{1}{n}\|D\|^{-1}_+\sum\limits_{i=1}^n
D({\bf x}_i)\Delta S_i\ge -c n^{-\frac{1}{4}+\epsilon}-\gamma,
\label{asympt-1an}
\end{eqnarray}
where $c$ is a positive constant.

By (\ref{asympt-1a}), for any $n$ and $\gamma>0$,
the inequality (\ref{asympt-1an}) fails with probability $2e^{-n\gamma^2}$.
Since given $\gamma>0$ the series $\sum\limits_{n=1}^\infty e^{-n\gamma^2}$
converges, by Borel--Cantelli lemma, for any $\gamma>0$
the inequality (\ref{asympt-1an}) can be violated
not more than for a finite number of different $n$. Hence, the event
(\ref{asympt-1f}) holds almost surely.
This completes the proof of Theorem~\ref{main-r1a}. $\triangle$

Theorem~\ref{main-r1a} can be rewritten for
the strategy $\tilde M_i^L=L\tilde M_i$ and for the class of
stationary strategies $D\in\cal F$ with bounded norm
$\|D\|_\infty\le L$, where $L$ is an arbitrary positive integer
number.

We present the following evident corollary for $\tilde M_i^L$.
\begin{corollary}\label{analo-1}
Given a positive integer number $L$, for any
stationary trading strategy $D\in\cal F$ such that
$\|D\|_\infty\le L$,
\begin{eqnarray*}
\liminf\limits_{n\to\infty}\left(\frac{1}{n}\sum\limits_{i=1}^n
\tilde M_i^L
\Delta S_i-\frac{1}{n}
\sum\limits_{i=1}^n D({\bf x}_i)\Delta S_i\right)\ge 0
\end{eqnarray*}
holds almost surely.

For any $\epsilon>0$, this trading strategy $\tilde M_i^L$
can be tuned such that for any $n$ and $\delta>0$,
with probability at least $1-\delta$, for
all nonnegative $D\in\cal F$ such that $\|D\|_\infty\le L$,
\begin{eqnarray*}                               	
\sum\limits_{i=1}^n \tilde M_i^L
\Delta S_i\ge\sum\limits_{i=1}^n D({\bf x}_i)\Delta S_i-
\nonumber
\\
-52L
(c^2_{{\cal F}}+1)^{\frac{1}{4}}
n^{\frac{3}{4}+\epsilon}-\|D\|_{{\cal F}}\sqrt{(c^2_{{\cal F}}+1)n}-
2L\sqrt{\frac{n}{2}\ln\frac{2}{\delta}}.
\end{eqnarray*}
\end{corollary}

\section{Universal consistency}\label{universal-1}

Using a universal kernel and the corresponding canonical universal RKHS,
we can extend our asymptotic results for all continuous
stationary trading strategies.

An RKHS $\cal F$ on $X$ is universal if $X$ is a compact metric space
and every continuous function $f$ on $X$ can be arbitrarily well approximated
in the metric $\|\cdot\|_\infty$ by a function from $\cal F$:
for any $\epsilon>0$ there exists $D\in\cal F$ such that
$$
\sup\limits_{x\in X}|f(x)-D(x)|\le\epsilon
$$
(see~\citealt{Ste2001}, Definition 4).

We use $X=[0,1]$. The Sobolev space ${\cal F}=H^1([0,1])$ defined in
Section~\ref{prem-1} is the universal RKHS
(see~\citealt{Ste2001}, \citealt{Vov2005a}).

We call a randomized trading strategy $\tilde M_i$ universally consistent
if for any continuous function $f$ with probability one
\begin{eqnarray}                               	
\liminf\limits_{n\to\infty}\left(\frac{1}{n}\sum\limits_{i=1}^n
\tilde M_i(S_i-S_{i-1})-\frac{1}{n}\|f\|_+^{-1}\sum\limits_{i=1}^n
f({\bf x}_i)(S_i-S_{i-1})\right)\ge 0.
\label{asymp-opt-1}
\end{eqnarray}
This definition is similar to~\citet{Vov2005a} definition of
a universally consistent prediction strategy.

The existence of the universal RKHS on $[0,1]$ implies the following
\begin{theorem}\label{iniversal-consist-1}
An algorithm for computing forecasts $p_i$ and a sequential method of
randomization can be constructed which performs at least as well as any
continuous trading strategy $f$:
\begin{eqnarray}
\liminf\limits_{n\to\infty}\left(\frac{1}{n}\sum\limits_{i=1}^n
\tilde M_i\Delta S_i-\frac{1}{n}\|f\|^{-1}_+
\sum\limits_{i=1}^n f({\bf x}_i)\Delta S_i\right)\ge 0
\label{asympt-1fuut}
\end{eqnarray}
holds almost surely with respect to a probability distribution
generated by the corresponding sequential randomization.
\end{theorem}
This result directly follows from the possibility to approximate
arbitrarily close any continuous function $f$ on $[0,1]$ by a function $D$
from the universal RKHS $\cal F$:
for any continuous function $f$ and for any $0<\epsilon<1$
take a $D\in\cal F$ such that
$\|f-D\|_\infty<\frac{1}{5}\epsilon\|f\|_+$. Then
\begin{eqnarray}
\liminf\limits_{n\to\infty}\left(\frac{1}{n}\sum\limits_{i=1}^n
\tilde M_i
\Delta S_i-\frac{1}{n}\|f\|^{-1}_+
\sum\limits_{i=1}^n f({\bf x}_i)\Delta S_i\right)+\epsilon\ge
\nonumber
\\
\ge\liminf\limits_{n\to\infty}\left(\frac{1}{n}\sum\limits_{i=1}^n
\tilde M_i
\Delta S_i-\frac{1}{n}\|D\|^{-1}_+
\sum\limits_{i=1}^n D({\bf x}_i)\Delta S_i\right)\ge 0.
\label{asympt-1fuuw}
\end{eqnarray}
Since (\ref{asympt-1fuuw}) holds for each $\epsilon>0$,
(\ref{asympt-1fuut}) is valid.

The property of universal consistency is
asymptotic and does not tell
us anything about finite data sequences: we cannot obtain the
convergence bounds like (\ref{asympt-1a}) and (\ref{asympt-1abf})
which holds for stationary strategies from RKHS.

\section{Competing with discontinuous trading strategies}\label{universal-2}

The trading strategy $\tilde M_i$ defined in Section~\ref{sec-2}
performs at least as well as any stationary trading strategy $D({\bf x})$
(up to some regret) even if the future price $S_i$ of the stock is known to $D$
as a side information contained in ${\bf x}_i$.
Theorems~\ref{main-r1}~and~\ref{main-r1a} are also valid in
this case.

This impressive efficiency of the trading strategy $\tilde M_i$ can be explained
by the restrictive power of continuous functions.
A weak point of {\it Trader D} is that a set of his strategies is limited by $\cal F$.
A continuous stationary trading strategy $D$ cannot respond sufficiently
quickly to information about changes of the value of a future price $S_i$.
the optimal trading strategy $\tilde M_i$,
is a discontinuous function, though it is applied to the random variables.

A positive argument in favor of the requirement of continuity of $D$ is that
it is natural to compete only with computable trading strategies,
and continuity is often regarded as a necessary condition for computability
(Brouwer's ``continuity principle'').

If $D$ is allowed to be discontinuous, we cannot prove (\ref{asympt-1})
and (\ref{asympt-1f}) in general case. We demonstrate the weakness
of discontinuous $D$ in Theorem~\ref{discontinuous-1} below.

Let an arbitrary randomizing trading strategy be given
that is a sequence of random variables $\tilde M_i$, $i=1,2,\dots$.
We suppose that they are independent like random variables that form
the universal trading strategy defined in Section~\ref{sec-2}.

A stationary trading strategy $D({\bf x})$ is called {\it decision rule}
if its range is finite. Decision rule is binary if it takes only two
values.

Consider the protocol of trading game presented on Fig~\ref{fig-4}
with two players and with signals that are probabilities:
$$
{\bf x}_i=P\{\tilde M_i>0\}
$$
for $i=1,2,\dots$.

Define a sequence of stock prices: $S_0=1/2$ and for $1\le i\le 1$
\[
S_i=
\left\{
    \begin{array}{l}
      S_{i-1}-2^{-(i+1)} \mbox{ if } {\bf x}_i>\frac{1}{2}
    \\
      S_{i-1}+2^{-(i+1)} \mbox{ otherwise}.
    \end{array}
  \right.
\]
By definition $S_i>0$ for all $i$.

Define the binary decision rule $D$:
\[
D(y)=
\left\{
    \begin{array}{l}
      -1 \mbox{ if } y>\frac{1}{2}
    \\
      1 \mbox{ otherwise},
    \end{array}
  \right.
\]
where $y\in [0,1]$.


\begin{theorem}\label{discontinuous-1}
Let $\tilde M_i$ an arbitrary randomizing trading strategy such that
$|\tilde M_i|\le 1$ for all $i$.

Then, with probability one,
\begin{eqnarray}
\limsup\limits_{n\to\infty}\left(\frac{1}{n}
\sum\limits_{i=1}^n\tilde M_i\Delta S_i-
\frac{1}{2}\frac{1}{n}\sum\limits_{i=1}^n D({\bf x}_i)\Delta S_i\right)\le 0,
\label{outp-1}
\end{eqnarray}
where $\Delta S_i=S_i-S_{i-1}$.
Inequality (\ref{outp-1}) means that trading strategy $D$ outperforms
$\tilde M_i$ twice.
\end{theorem}
{\it Proof}. We bound the conditional mathematical expectation of the random
variable $\tilde M_i$:
\begin{eqnarray}
E(\tilde M_i)=\int\limits_{\tilde M_i>0}\tilde M_idP+
\int\limits_{\tilde M_i\le 0}\tilde M_idP\le
P\{\tilde M_i>0\}={\bf x}_i.
\label{m--bound-1}
\\
E(\tilde M_i)\ge
-P\{\tilde M_i\le 0\}={\bf x}_i-1.
\label{m--bound-2}
\end{eqnarray}

If ${\bf x}_i>\frac{1}{2}$ then
$E(\tilde M_i)\ge -\frac{1}{2}$
by (\ref{m--bound-2}), $\Delta S_i=-2^{-(i+1)}$, and $D({\bf x}_i)=-1$
by definition.

If ${\bf x}_i\le\frac{1}{2}$ then
$E(\tilde M_i)\le\frac{1}{2}$
by (\ref{m--bound-1}), $\Delta S_i=2^{-(i+1)}$, and $D({\bf x}_i)=1$
by definition.
We have for any $n$,
\begin{eqnarray}
E\left(\sum\limits_{i=1}^n \tilde M_i\Delta S_i\right)=
\sum\limits_{i=1}^n E(\tilde M_i)\Delta S_i=
\nonumber
\\
=\sum\limits_{{\bf x}_i>\frac{1}{2}}^n
E(\tilde M_i)\Delta S_i+
\sum\limits_{{\bf x}_i\le\frac{1}{2}}^n
E(\tilde M_i)\Delta S_i\le\frac{1}{2}
\sum\limits_{i=1}^n 2^{-(i+1)}=\frac{1}{4}.
\end{eqnarray}
Also,
\begin{eqnarray}
\sum\limits_{i=1}^n D({\bf x}_i)\Delta S_i=
\sum\limits_{{\bf x}_i>\frac{1}{2}}^n
D({\bf x}_i)\Delta S_i+
\sum\limits_{{\bf x}_i\le\frac{1}{2}}^n D({\bf x}_i)\Delta S_i=
\sum\limits_{i=1}^n 2^{-(i+1)}=\frac{1}{2}.
\end{eqnarray}
By the law of large numbers, with probability 1:
\begin{eqnarray}
\frac{1}{n}\sum\limits_{i=1}^n (\tilde M_i-E(\tilde M_i)\to 0
\label{lawoflarge}
\end{eqnarray}
as $n\to\infty$. From this (\ref{outp-1}) follows.
Theorem is proved. $\triangle$

Theorem~\ref{discontinuous-1} is valid in a more general setting where
random variable $\tilde M_i$, $i=1,2,\dots$ are be dependent. In this
case we have to use signals that
are random variables representing conditional probabilities:
${\bf x}_i=P\{\tilde M_i>0|\tilde M_1,\dots ,\tilde M_{i-1}\}$.
The proof of Theorem~\ref{discontinuous-1} is almost the same
but we have to consider conditional mathematical expectation
$E(\tilde M_i|\tilde M_1,\dots ,\tilde M_{i-1})$
in (\ref{m--bound-1}) and in what follows.
\footnote
{
In general case the law of large numbers (\ref{lawoflarge})
is a corollary of Azuma--Hoeffding inequality be
applied for martingale-differences $V_i=
\tilde M_i-E(\tilde M_i|\tilde M_1,\dots ,\tilde M_{i-1})$ (see~\cite{Ces2006}).
}

The discontinuous trading strategy $D$ defined in Theorem~\ref{discontinuous-1}
is unstable under small changes of the signal ${\bf x}_i$.
In the next theorem, we show that if we randomly round the
signal ${\bf\tilde x}_i$ then our universal trading strategy $\tilde M_i$
(and $\tilde M^1_i$), performs at least as well as $D$.

Consider the protocol of trading game with two players
and a side information ${\bf x}_i\in [0,1]$ (see Fig~\ref{fig-4}).

We specify the information vector using by our universal strategy
$\tilde M_i$ to be $\bar x_i=(S_{i-1},{\bf x}_i)$,
where $S_{i-1}$ is the past price of the stock
and ${\bf x}_i$ is the signal at step $i$. The universal trading strategy
$\tilde M_i$ uses the sequential method of randomization defined in
Section~\ref{prem-1} to perform a randomized forecast $\tilde p_i$ and
a randomized information vector
$\tilde x_i=(\tilde S_{i-1},{\bf\tilde x}_i)$.

The strategy of {\it Trader M} is the same as before:
\[
\tilde M_i=
\left\{
    \begin{array}{l}
      1 \mbox{ if } \tilde p_i>\tilde S_{i-1},
    \\
      -1 \mbox{ otherwise, }
    \end{array}
  \right.
\]
except that it uses a slightly different randomization.

\begin{theorem}\label{main-r1des}
An algorithm for computing forecasts and a sequential
method of randomization
can be constructed such that for any decision rule $D$
\begin{eqnarray}
\liminf\limits_{n\to\infty}\left(\frac{1}{n}\sum\limits_{i=1}^n
\tilde M_i
\Delta S_i-\frac{1}{n}\|D\|^{-1}_+
\sum\limits_{i=1}^n D({\bf\tilde x}_i)\Delta S_i\right)\ge 0
\label{asympt-1des}
\end{eqnarray}
holds almost surely with respect to a probability distribution
generated by the corresponding sequential randomization.

Moreover, for any $\epsilon>0$ this trading strategy $\tilde M_i$
can be tuned such that for any $\delta>0$ and $n$,
with probability at least $1-\delta$, for
all nonnegative decision rule $D\in\cal F$,
\begin{eqnarray}                               	
\sum\limits_{i=1}^n
\tilde M_i\Delta S_i\ge\|D\|^{-1}_+\sum\limits_{i=1}^n
D({\bf\tilde x}_i)\Delta S_i-
\nonumber
\\
-25(1+m) n^{\frac{4}{5}+\epsilon}-
(1+m)\sqrt{\frac{n}{2}\ln\frac{2m}{\delta}},
\label{asympt-1ades}
\end{eqnarray}
where $m$ is the cardinality of the range of $D$.
\end{theorem}
{\it Proof}. For simplicity, we give the proof for the case of nonnegative
decision rule and the randomized strategy $M^1_i$.
The case of arbitrary decision rule $D$ and strategy $\tilde M_i$
is considered similarly.

We apply Theorem~\ref{univ-1b} to zero kernel $K_2({\bf x},{\bf x'})=0$
with $c_{{\cal F}}=0$ and to the information vector
$\bar x_i=(S_{i-1},{\bf x}_i)$, $k=2$.

Recall that $\epsilon>0$ and $M=\lceil 2/\epsilon\rceil$.
At any step $i$ we compute the deterministic forecast $p_i$ defined
in Theorem~\ref{univ-1b} (Section~\ref{gam-1a}) and its randomization
to $\tilde p_i$
using parameters $\Delta=\Delta_s=2(3/4)^{2/5}(s+M)^{-\frac{M}{5}}$
and $n_s=(s+M)^M$, where $n_s\le i<n_{s+1}$.

The following upper bound is valid:
\begin{eqnarray}
\left|\|D\|^{-1}_+\sum\limits_{i=1}^n D({\bf\tilde x}_i)
(\tilde S_{i-1}-S_{i-1})\right|\le\sum\limits_{t=0}^s(n_{t+1}-n_t)\Delta_t\le
5n_s^{\frac{4}{5}},
\label{esima-1i}
\end{eqnarray}
where $n_s\le n< n_{s+1}$.

Let $D({\bf x})$ be an arbitrary nonnegative decision rule.
Let $\tilde M_i^1$ be the randomized trading strategy
defined in Section~\ref{sec-2}.
We use abbreviations:
\begin{eqnarray}
\nu_1(n)=5n^{\frac{4}{5}},
\label{hoeff-1i}
\\
\nu_2(n)=20n^{\frac{4}{5}+\epsilon}+\sqrt{\frac{n}{2}\ln\frac{2m}{\delta}}.
\label{call-1abui}
\end{eqnarray}
All sums below are for $i=1,\dots n$. By definition
$0\le D({\bf\tilde x}_i)\le\|D\|_+$ for all ${\bf x}_i\in [0,1]$.

Let $d_1,\dots,d_m$ be all values of $D$.
Define
$$
R_j=\{(p,y,{\bf x}):0\le p,y\le 1, D({\bf x})=d_j\},
$$
where $1\le j\le m$. Let $I_{R_j}$ be the characteristic
function of the set $R_j$.

Let $\delta>0$ and $n$ be given. Then, with probability $1-\delta$,
the following chain of equalities and inequalities is valid:
\begin{eqnarray}
\sum\limits_{i=1}^n \tilde M_i^1
(S_i-S_{i-1})=
\sum\limits_{\tilde p_i>\tilde S_{i-1}}(S_i-S_{i-1})=
\nonumber
\\
=\sum\limits_{\tilde p_i>\tilde S_{i-1}}(S_i-\tilde p_i)+
\sum\limits_{\tilde p_i>\tilde S_{i-1}}(\tilde p_i-\tilde S_{i-1})+
\sum\limits_{\tilde p_i>\tilde S_{i-1}}(\tilde S_{i-1}-S_{i-1})\ge
\label{azu-1i}
\\
\ge\sum\limits_{\tilde p_i>\tilde S_{i-1}}
(\tilde p_i-\tilde S_{i-1})-\nu_1(n)-\nu_2(n)\ge
\label{azu-2i}
\\
\ge\|D\|^{-1}_+\sum\limits_{i=1}^n D({\bf\tilde x}_i)
(\tilde p_i-\tilde S_{i-1})-\nu_1(n)-\nu_2(n)=
\label{azu-2hi}
\\
=\|D\|^{-1}_+\sum\limits_{i=1}^n D({\bf\tilde x}_i)
(S_i-S_{i-1})-\nu_1(n)-\nu_2(n)-
\nonumber
\\
-\|D\|^{-1}_+\sum\limits_{i=1}^n D({\bf\tilde x}_i)
(\tilde S_{i-1}-S_{i-1})-
\|D\|^{-1}_+\sum\limits_{i=1}^n D({\bf\tilde x}_i)(S_i-\tilde p_i)\ge
\label{azu-2hii}
\\
\ge\|D\|^{-1}_+\sum\limits_{i=1}^n D({\bf\tilde x}_i)
(S_i-S_{i-1})-
(1+m)\nu_1(n)-(1+m)\nu_2(n).
\label{azu-2hiii}
\end{eqnarray}
In change from (\ref{azu-1i}) to (\ref{azu-2i}) and in change
from (\ref{azu-2hii}) to (\ref{azu-2hiii}) we have used
the inequality (\ref{esima-1i}). In change
from (\ref{azu-2hii}) to (\ref{azu-2hiii}) we have used also
Theorem~\ref{univ-1b}, where $k=2$, and, with probability $1-\delta$,
\begin{eqnarray*}
\left|\sum\limits_{i=1}^n D({\bf\tilde x}_i)(S_i-\tilde p_i)\right|=
\left|\sum\limits_{j=1}^m d_j
\sum\limits_{i=1}^n I_{R_j}({\bf\tilde x}_i)
(S_i-\tilde p_i)\right|\le m\|D\|_+\nu_2(n).
\end{eqnarray*}

The inequality (\ref{asympt-1des}) follows from (\ref{asympt-1ades}).
Theorem~\ref{main-r1des} is proved. $\triangle$

\begin{figure}[t]
\centering\includegraphics[height=60mm,width=155mm,clip]{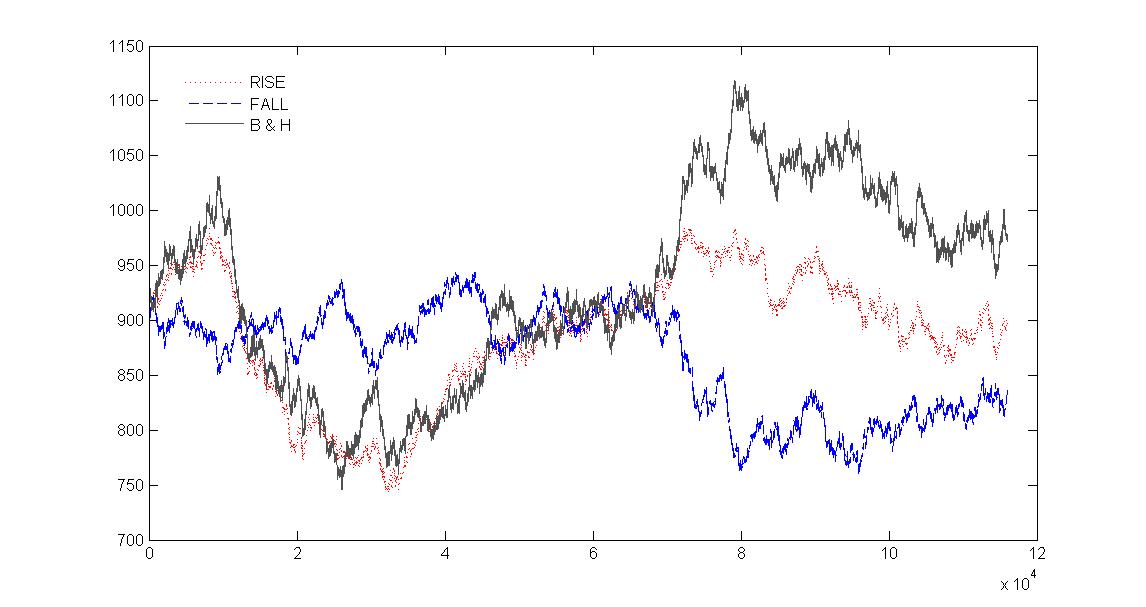}
         \caption{Evolution of capitals of three trading strategies
for the period 26.03.10--25.03.11:
Buy and Hold -- solid line, UN going long -- dotted line,
UN going short -- dashed line.
One run of trading is performed with a simulated stock TEST
(see Table~\ref{table-1})}\label{fig-5}
                  \end{figure}

\begin{figure}[t]
\centering\includegraphics[height=60mm,width=165mm,clip]{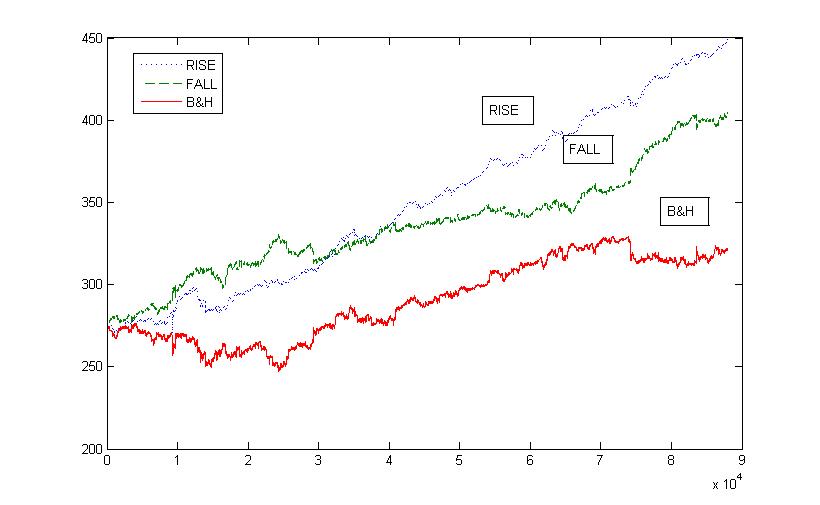}
          \caption{Evolution of capitals of three trading strategies
for the period 26.03.10--25.03.11:
Buy and Hold -- solid line, UN going long -- dotted line,
UN going short -- dashed line.
One run of trading is performed with the stock KOCO
(see also Table~\ref{table-1})}\label{fig-6}
                  \end{figure}



\section{Numerical experiments}\label{app-1}

{\bf Computer technology}.
In the numerical experiments, we have used historical data in the
form of per minute time series of prices of arbitrarily chosen stocks.


Two types of kernel functions
were used as the smooth approximations of the
combined kernel
${\cal K}(p,p_n,x,x_i)=K_1(p,x_n,p_i,\bar x_i)+K_2({\bf x}_n,{\bf x}_i)$ from
the sum (\ref{kernel-comb-1}): (i) ${\cal K}(p,p_n)=\cos((\pi(p-p_n)/2)$,
(ii) ${\cal K}(p,p_n,x,x_i)=\exp(c(p-p_n)+c'(x-x_i))$,
where $c,c'$ are positive constants.

\begin{figure}

\begin{picture}(100,100)
\put(0,30){\line(1,0){100}}
\put(40,30){$\rightarrow$}
\put(100,17){$\downarrow$}
\put(130,10){$\rightarrow$}

\put(60,10){\line(1,0){100}}
\put(160,17){$\uparrow$}
\put(190,30){$\rightarrow$}

\put(120,30){\line(1,0){100}}
\put(220,17){$\downarrow$}
\put(250,10){$\rightarrow$}

\put(240,30){\line(1,0){100}}
\put(280,17){$\uparrow$}
\put(310,30){$\rightarrow$}

\put(180,10){\line(1,0){100}}
\put(340,17){$\downarrow$}
\put(370,10){$\rightarrow$}

\put(300,10){\line(1,0){100}}

\end{picture}

\caption{Scheme of parallel computations}\label{fig-7}
\end{figure}
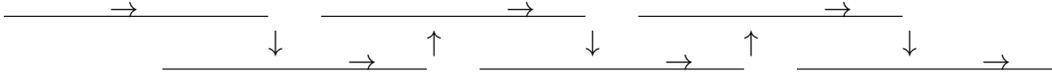

In any short-term trading algorithm, the time characteristics are crucial.
The greatest time cost is associated with the calculation
of sums (\ref{kernel-comb-1}) and finding the roots of this equation.
The performed experiments show that the computation time for one point of
the forecast increases linearly with increasing length of
history.
To provide one point of time, predicting within 1 - 3 seconds
of CPU time, the length of the series was limited
up to 5000 points. For series of length greater than 5000 points,
``a chain'' method of forecasting was used.
Two processes working on overlapping intervals of time series
are performed at the same time (see Fig~\ref{fig-7}).

Let $L_{{\rm max}}$ be the chain length,
and $L_{{\rm shift}}$ be the value of time shift, where
$L_{{\rm shift}}<L_{{\rm max}}$. In any process, the first $L_{{\rm shift}}$
time-points are used only for scaling prices
and preliminary learning of the forecasting algorithm.
The trading is not performed
at first $L_{{\rm shift}}$ time-points of the series.

When a regular process terminates we switch to the time-point
$L_{{\rm shift}}+1$ of the next process.
The results of parallel computing are accumulated into a single overall
forecasting series. We take $L_{{\rm max}}=5000$ and $L_{{\rm shift}}=2000$.

The prices of a stock are scaled such that $S_i\in [0,1]$ for all $i$.
The scaling is performed for time series of each process separately. The first
$L_{{\rm shift}}$ time points of any process are used for computing
a scaling constant. Prices are scaled as follows:
$$
S_i=\frac{\hat S_i}{c\max\limits_{1\le j\le L_{{\rm shift}}}\hat S_j},
$$
where $1\le i\le L_{{\rm max}}$ and $\hat S_i$ are real prices of the stock.
We set $c=14$.

The forecasting algorithm is performed for the scaled prices $S_i$,
where $L_{{\rm shift}}+1\le i\le L_{{\rm max}}$.

We implement this computer technology for two forecasting algorithms:
the universal strategy constructed in Section~\ref{gam-1a} (UN--model)
and Autoregressive Moving Average algorithm
(ARMA--model) (see~\citealt{JPJ2011}).
\footnote
{
See also the State Space Models Toolbox for
MATLAB:\\
{\it http://sourceforge.net/projects/ssmodels/}.
}

\begin{table}[t]
\caption{Universal trading}
\label{table-1}
\vskip 0.15in
\begin{center}
\begin{small}
\begin{sc}
\begin{tabular}{|l|l|l|l|l|l|l|l|l|l|l|}
\hline
       & Buy\&  &  UN & UN   & ARMA  & ARMA     \\
Ticker & Hold   &  going long  & going short    &going long   & going short    \\
       & Profit \%& Profit \%& Profit \% &Profit \%& Profit \% \\
\hline
TEST   & 6.85   & -1.39  &-8.19    & 9.88   &3.08    \\
\hline
AT-T   & 7.71   & 137.40 & 129.70  & 30.73   &23.02   \\
\hline
CTGR   & 15.04  & 1594.34&1579.34  &1167.22  &1152.53  \\
\hline
KOCO   & 16.55  & 62.66  & 46.15   &2.90    &-13.61    \\
\hline
GOOG   & 10.25  & 114.85 & 104.62  &12.85    &2.62    \\
\hline
InBM   & 24.28  & 85.38  & 61.09   &29.31    &5.02    \\
\hline
INTL   & 4.29   & 111.70 & 107.50  &25.86    &21.66    \\
\hline
MSD& 10.71 & 58.32 & 47.60    &18.66    &7.95    \\
\hline
US1.AMT& 22.01  & 22.74 & 0.77     &28.46    &6.49     \\
\hline
US1.IP  &2.40   & 19.83 & 17.47    &9.36    &7.00   \\
\hline
US2.BRCM& 25.30  & 53.62 & 28.28    &20.06    &-5.27    \\
\hline
US2.FSLR&40.15  & 143.92 & 103.61  &-9.86    &-50.16   \\
\hline
SIBN   & -6.54  & 732.87 & 739.33  & 357.74  &364.20   \\
\hline
GAZP   & 22.75  & 101.20 & 78.45   & 31.75   &9.00    \\
\hline
LKOH   & 19.39  & 261.84 & 242.45  & 87.08   &67.68  \\
\hline
MTSI   & -1.61  & 669.16 & 670.68  & 326.12  &327.64   \\
\hline
ROSN   & 9.69   & 188.89 & 179.12  &34.40    &24.63    \\
\hline
SBER   & 14.21  & 108.97 & 94.90   &37.53    &23.46    \\
\hline
\end{tabular}%
\end{sc}
\end{small}
\end{center}
\vskip -0.1in
\end{table}

{\bf Results of numerical experiments}.
In the numerical experiments, we have used historical data in
form of per minute time series of prices of arbitrarily chosen
17 stocks (11 US stocks, and 6 Russian stocks) and of one simulated stock
TEST. Data has been downloaded from FINAM site: \url{www.finam.ru}.
Number of trading points in each game is N=88000--116000 min.
(From March 26 2010 to March 25 2011).

The artificial stock TEST is simulated as $S_i=S_{i-1}+\xi_i$, $i=1,2,\dots, N$,
where $\xi_i$ is the Gaussian random variable with mean 0 and a
variance equal to the variance of the scaled GAZP stock.

We implement the trading strategy defined in Section~\ref{sec-2}.

Two series of numerical experiments were performed.

In the first series, we use the trading strategy $\tilde M_i$ studied in
Theorem~\ref{main-r1a}. At each step, starting from initial capital
${\cal K}_0^R={\cal K}_0^F={\cal K}_0=KS_0$,
where $S_0$ is the price of a stock at the first
time point, this strategy performs going long or for going short with $K$
shares of the stock. We take $K=5$ in our experiments.
In case of going long, the capital changes at any step $i$ as
${\cal K}_i^R={\cal K}_{i-1}^R+K(S_i-S_{i-1})$ if
$\tilde p_i>\tilde S_{i-1}$ and ${\cal K}_i^R={\cal K}_{i-1}^R$ otherwise.
In case of dealing for a fall ${\cal K}_i^F={\cal K}_{i-1}^F-K(S_i-S_{i-1})$
if $\tilde p_i\le\tilde S_{i-1}$ and ${\cal K}_i^F={\cal K}_{i-1}^F$
otherwise, where $i=1,2,\dots, N$.

Results of numerical experiments
are shown in Table~\ref{table-1}. In the first column,
stocks ticker symbols are shown. The second column contains the
profit of Buy-and-Hold trading strategy. By this strategy, we buy a
holding of shares using capital ${\cal K}_0$ and sell them for ${\cal K}_N$
at the end of the trading period.

\begin{table}[t]
\caption{Defensive trading}
\label{table-2}
\vskip 0.15in
\begin{center}
\begin{small}
\begin{sc}
\begin{tabular}{|l|l|l|l|l|l|l|l|l|l|l|}
\hline
       & Buy\&  & UN      &UN     &ARMA    &ARMA   &UN &ARMA & UN   & ARMA \\
Ticker & hold   &Profit&Profit&Profit&Profit&   &      &      &      \\
       & \%     &\%&-0.01\% &\%     &-0.01\%            &In&In  & D& D \\
\hline
TEST   & 6.85  & 3.58  &-80.93 & 3.58 & -80.90& 0.232  & 0.163 & 1.453 & 1.890 \\
\hline
AT-T   & 7.71 & 69.01   &-79.19 & 29.86  &-79.19& 0.218 & 0.205 & 1.611 & 1.576 \\
\hline
CTGR   & 15.04 & 1030.12 &658.13 & 937.46 &540.18& 0.238 & 0.253 & 1.654 & 1.479  \\
\hline
KOCO   & 16.55 & 36.47  &-78.62 & 15.69 &-78.55& 0.216 & 0.198 & 1.604 & 1.502 \\
\hline
GOOG   & 10.25 & 46.54  &-80.57 & 3.53 &-82.68 & 0.231 & 0.211 & 1.462 & 1.474 \\
\hline
InBM   & 24.28 & 54.79  &-78.53 & 34.66  &-78.10& 0.219 & 0.187 & 1.514 & 1.517 \\
\hline
INTL   & 4.29 & 43.06   &-76.60 & 5.63  &-76.28& 0.220 & 0.179 & 1.630 & 1.585 \\
\hline
MCD   & 10.71 & 34.22   &-78.56 & 19.21 &-78.41& 0.222 & 0.190 & 1.571 & 1.876 \\
\hline
AMT  & 22.01 & 16.47   &-77.01 & 24.04  &-77.09& 0.212 & 0.183 & 1.654 & 1.758 \\
\hline
IP   & 2.40 & 4.45   &-82.78 & -14.79  &-81.06& 0.213 & 0.181 & 1.657 & 1.760 \\
\hline
BRCM & 25.30 & 11.40   &-80.47 & 23.98  &-76.10& 0.216 & 0.172 & 1.585 & 1.876 \\
\hline
FLSR & 40.15 & 21.02   &-80.04 & -27.50  &-80.03& 0.227 & 0.196 & 1.499 & 1.506 \\
\hline
SIBN   & -6.54 & 600.62 &249.87 & 287.48 &-58.55& 0.169 & 0.179 & 2.460 & 2.292 \\
\hline
GAZP   & 22.75 & 51.29  &-82.04 & 4.34  &-82.16 & 0.224 & 0.210 & 1.539 & 1.526 \\
\hline
LKOH   & 19.39 & 149.03  &-79.91 & 46.44  &-80.62 & 0.230 & 0.244 & 1.527 & 1.501 \\
\hline
MTSI   & -1.61 & 482.83 &79.23  & 275.13 &-69.36& 0.188 & 0.195 & 2.174 & 1.959 \\
\hline
ROSN   & 9.69 & 101.15  &-83.14 & -0.53  &-83.54 & 0.228 & 0.240 & 1.549 & 1.499 \\
\hline
SBER   & 14.21 & 51.56  &-82.52 & -14.47  &-82.73& 0.225 & 0.196 & 1.559 & 1.674 \\
\hline
\end{tabular}%
\end{sc}
\end{small}
\end{center}
\vskip -0.1in
\end{table}

In the 3th and 4th columns, results of one run of trading based on
the universal randomized
forecasting strategy (UN) are shown. In the 3th column,
a relative return, percentagewise, to the initial capital
$\frac{{\cal K}_N-{\cal K}_0}{{\cal K}_0}100\%$ is shown for going long,
in the 4th column, the same relative return
is shown for going short,
In the 5th and 6th columns, the same results are shown for
trading using ARMA forecasts.

It was found that ${\cal K}_i>0$ for $i=1,2,\dots, N$, i.e.,
we never incur debt in our experiments (with an exception of TEST stock).

Results presented in Table~\ref{table-2} show that trading based on UN model of
forecasting performs at least as well as the trading based on ARMA forecasting model and
essentially outperforms it for some stocks.

The second series of experiments is closer to a real short-term trading.
The trading strategy has a defence guarantee. Starting with the same initial
capital ${\cal K}_0=KS_0$, where $S_0$ is the initial price of a stock
and $K=5$, we perform going long using ``a defensive'' trading strategy.
At any step $i$, our working capital is
${\cal L}_{i-1}=\min\{{\cal K}_0,{\cal K}_{i-1}\}$. Using this capital,
we buy $M_i={\cal L}_{i-1}/S_{i-1}$ shares of the stock at the
beginning of any step $i$, if ${\cal L}_{i-1}>0$, and stop trading otherwise:
$M_i=0$. We update the cumulative
capital at the end of each step: ${\cal K}_i={\cal K}_{i-1}+M_i(S_i-S_{i-1})$.
Thereby, we can set aside the extra income.

Results of second series of numerical experiments are shown in
Table~\ref{table-2}. In the first column, stocks ticker symbols
are shown. The second column contains the relative return of
Buy-and-Hold trading strategy.
In the next pair of columns marked ``UN'', relative returns
of one run of randomized trading, percentagewise,
for the initial capital are presented
for the case with no transaction costs and for the case where
transaction cost at the rate $0.01\%$ is subtracted.
We compute the forecast of a future stock price
by the method of calibration and defensive forecasting (UN)
presented in Theorem~\ref{univ-1b}.

The next two columns marked by ``ARMA'' are similar,
with the exception that the ARMA forecasting model is used for
computing forecasts. The frequencies of market entry steps
$i$, where $\tilde p_i>\tilde S_{i-1}$,
are given in the next two columns marked ``In'' (for UN and ARMA).
We sell all shares of a stock at step $i$ in case
$\tilde p_i\le\tilde S_{i-1}$.
The average time spent in the market is shown in the rest two columns
marked ``D'' (for UN and ARMA).

\section{Acknowledgement}

This research was partially supported by Russian foundation
for fundamental research: Grant 13-01-12447 and 13-01-0052.

\section{Conclusion}

Asymptotic calibration is an area of intensive research where several
algorithms for computing well-calibrated forecasts have been
developed. Several applications of well-calibrated forecasting
have been proposed (convergence to correlated equilibrium,
recovering unknown functional dependencies, predictions with expert
advice). We present a new application of the calibration
method.


We show that the universal trading strategy can be
constructed using the well-calibrated forecasts. We prove that this
strategy performs at least as well as any stationary trading strategy
presented by a rule from any RKHS with regret $O(n^{\frac{3}{4}})$.
Using the universal kernel, we prove that this strategy performs at
least as good as any stationary continuous trading strategy.

The obvious drawback of a universal
strategy is that it uses the high frequency trading,
which prevents it from practical applications in the presence of
transaction costs.

To construct the universal trading strategy, we
generalize Kakade and Foster's algorithm and combine it with
Vovk's DF--model for arbitrary RKHS. Using~\citet{Vov2006} theory of
defensive forecasting in Banach
spaces, these results can be generalized to these spaces.

Unlike in statistical theory, no stochastic assumptions are made about
the stock prices.

Numerical experiments show a positive return for all chosen
stocks, and for some of them we receive a positive return even when
transaction costs are subtracted. Results of this type can be
useful for technical analysis in finance.

\end{document}